\documentclass{article}

 \usepackage[preprint,nonatbib]{neurips_2026}

\usepackage{enumitem}
\usepackage[utf8]{inputenc} % allow utf-8 input
\usepackage[T1]{fontenc}    % use 8-bit T1 fonts
\usepackage{hyperref}       % hyperlinks
\usepackage{url}            % simple URL typesetting
\usepackage{booktabs}       % professional-quality tables
\usepackage{amsfonts}       % blackboard math symbols
\usepackage{nicefrac}       % compact symbols for 1/2, etc.
\usepackage{microtype}      % microtypography
\usepackage{xcolor}         % colors

\usepackage{mdframed}

\usepackage{amsmath} 
\usepackage{wrapfig}

\usepackage{graphicx}
\usepackage{subfigure}
\usepackage{booktabs} % for professional tables
\usepackage{multirow}
\usepackage{arydshln}

\usepackage{tcolorbox}
\usepackage{colortbl}

\usepackage[textsize=tiny]{todonotes}

\definecolor{ourblue}{rgb}{0.3,0.3,0.85}
\definecolor{darkgreen}{rgb}{0.0,0.6,0.0}
\definecolor{beaublue}{rgb}{0.9, 0.95, 0.9}
\definecolor{blackish}{rgb}{0.2, 0.2, 0.2}
\definecolor{grayish}{rgb}{0.86274,0.92980,0.98431}

\definecolor{KFblue}{HTML}{1F4E79}
\definecolor{KFbg}{HTML}{F2F7FF}

\definecolor{LightCyan}{rgb}{0.88,1,0.88}
\definecolor{LightRed}{rgb}{1,0.88,0.88}
\definecolor{LightBlue}{rgb}{0.86274,0.92980,0.98431}

\definecolor{ao(english)}{rgb}{0.0, 0.5, 0.0}
\title{Does Compression Preserve Uncertainty? A Unified Benchmark for Quantized and Sparse LLMs via Conformal Prediction}

\author{
Yujia Tong\textsuperscript{1}\thanks{Equal contribution.},
Yuxi Wang\textsuperscript{1}\footnotemark[1],
Yunyang Wan\textsuperscript{1},
Tian Zhang\textsuperscript{1},
Junhao Dong\textsuperscript{2}\thanks{Corresponding author.},
Jingling Yuan\textsuperscript{1}\footnotemark[2] \\
\textsuperscript{1} Wuhan University of Technology,
\textsuperscript{2}  Nanyang Technological University \\
{\tt\small \{tyjjjj, yjl\}@whut.edu.cn}, \tt\small junhao003@ntu.edu.sg \\
}

\begin{document}

\maketitle

\vspace{-10pt}
\begin{figure*}[htpb]
\centering
\includegraphics[width=\textwidth]{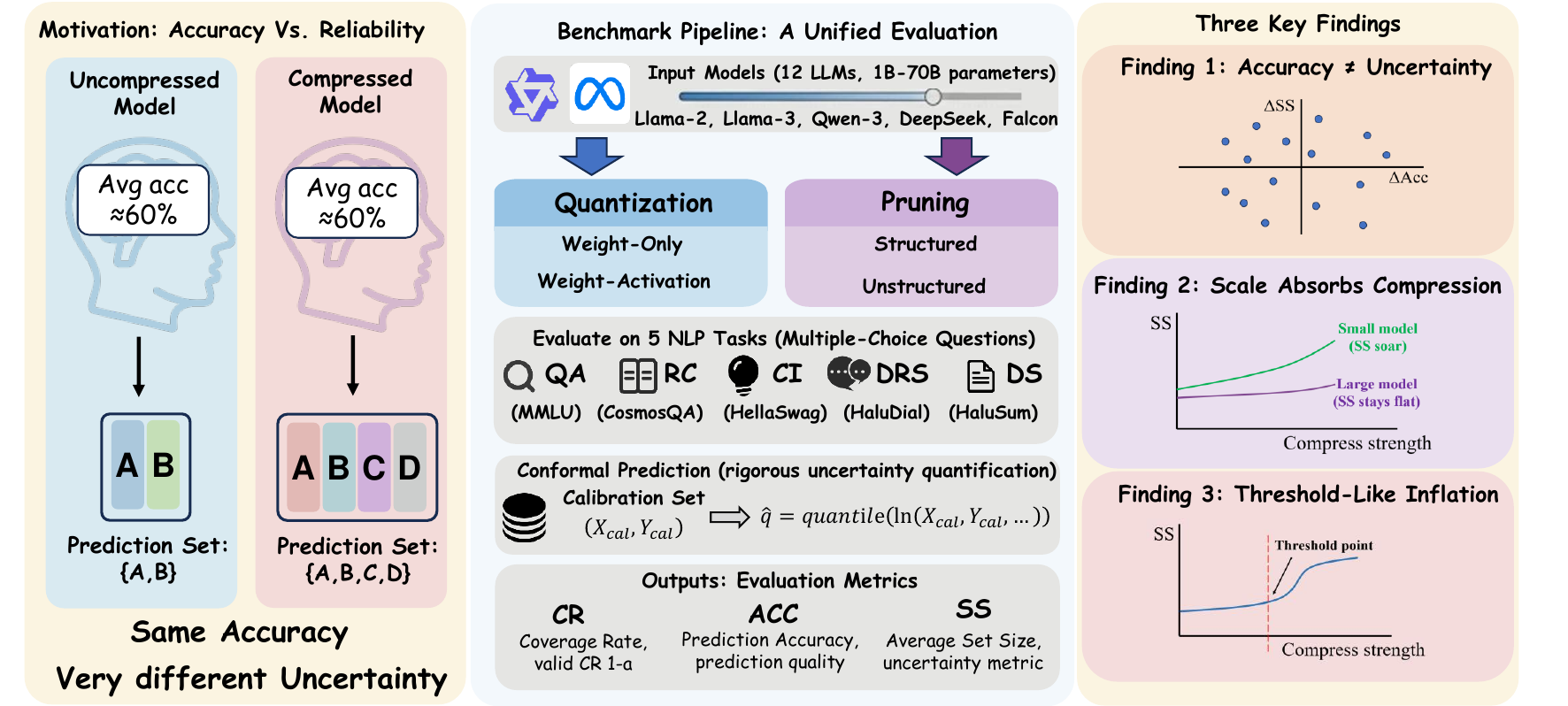}
\caption{Overview of our benchmark. Left: two models with the same accuracy can exhibit very different prediction set sizes, motivating uncertainty-aware evaluation beyond accuracy. Center: benchmark pipeline --- 12 LLMs (1B--70B) are compressed via four paradigms and evaluated on five NLP tasks using conformal prediction, yielding three complementary metrics. Right: three key findings revealed by our experiments.}
\label{Overview}
\end{figure*}

\begin{abstract}
Model compression techniques such as quantization and pruning are widely used to
reduce the deployment cost of large language models (LLMs), with existing
evaluations focusing almost exclusively on accuracy preservation. However, in
safety-critical applications, a model's ability to reliably quantify its own
uncertainty is equally important. We ask: does compression preserve this ability?
To answer this question, we benchmark 12 LLMs under various compression configurations
across five NLP tasks, using conformal prediction to provide a rigorous,
distribution-free measure of uncertainty. Our experiments reveal that:
\textit{(I) compression frequently decouples accuracy from uncertainty;
(II) larger models absorb compression-induced uncertainty far more effectively than smaller ones; and
(III) uncertainty inflation is often threshold-like rather than gradual.} These results suggest that accuracy-only evaluation is insufficient for assessing the deployment readiness of compressed LLMs, and that uncertainty-aware benchmarking should be a standard component of model compression pipelines.
\end{abstract}

\section{Introduction}
Large language models (LLMs)~\cite{grattafiori2024llama,achiam2023gpt,Dong_2026_CVPR,dong2026tug} have demonstrated remarkable capabilities across a wide spectrum of tasks, from question answering and commonsense reasoning to dialogue generation and document~\cite{li2026craft,li2026reason}. However, the substantial computational and memory requirements of these models pose significant barriers to their practical deployment, particularly in resource-constrained environments such as edge devices, real-time applications, and cost-sensitive production systems~\cite{tong2026sage}. To bridge this gap, model compression techniques~\cite{zhu2024survey,guo2025optimal} have emerged as indispensable tools for reducing the cost of LLM inference. Among them, quantization and pruning represent two dominant and complementary paradigms. Quantization methods~\cite{ashkboos2024quarot,lin2024awq,xiao2023smoothquant,dettmers2022gpt3,shao2024omniquant} reduce memory footprint by lowering the numerical precision of model weights, while pruning~\cite{ashkboos2024slicegpt,frantar2023sparsegpt,an2024fluctuation} approaches eliminate redundant parameters or structures to yield sparse, efficient models. While extensive efforts have been devoted to preserving task accuracy under compression, a critical yet underexplored question remains: do compressed LLMs still produce reliable uncertainty estimates? In high-stakes applications such as medical diagnosis, legal reasoning, and autonomous decision-making, a model's ability to quantify what it does not know~\cite{kadavath2022language} is as important as its predictive accuracy. Deploying a compressed model without understanding how compression affects this ability introduces risks that may undermine the very efficiency gains compression seeks to achieve.
 
To rigorously assess the uncertainty of LLMs, a quantification method that is both model-agnostic and statistically principled is needed. Conformal prediction~\cite{shafer2008tutorial,angelopoulos2023conformal} offers exactly such a framework. As a distribution-free approach to uncertainty quantification, it provides finite-sample coverage guarantees without assumptions on the underlying model or data distribution. By constructing prediction sets that provably contain the true label with a user-specified probability, conformal prediction yields a rigorous and interpretable measure of model reliability. Recent studies~\cite{ye2024benchmarking,quach2024conformal,cherian2024large,karimi2023quantifying} have applied this framework to evaluate the uncertainty of full-precision LLMs, demonstrating its effectiveness across tasks such as question answering, commonsense reasoning, and summarization. However, existing benchmarks focus exclusively on uncompressed models. It remains an open question whether compressed LLMs, despite preserving competitive accuracy, silently suffer from inflated uncertainty. To date, no existing work has systematically examined how quantization or pruning affects the uncertainty profile of LLMs under conformal prediction, nor has any study provided a unified comparison of these two compression paradigms from an uncertainty quantification perspective.
 
In this paper, we present the first comprehensive benchmark for evaluating the uncertainty of compressed LLMs through the lens of conformal prediction. Our benchmark encompasses two major compression paradigms, quantization and pruning, covering a broad spectrum of representative methods. We evaluate 12 LLMs spanning four model families and scales from 1B to 70B parameters under various compression configurations across five NLP tasks. Through systematic analysis, we uncover three findings that challenge the common assumption that accuracy preservation implies uncertainty preservation under compression. Our main contributions can be summarized as follows: 
 
\begin{itemize}[leftmargin=*]
 
    \item \textbf{A new evaluation perspective.} We are the first 
    to systematically evaluate the uncertainty of compressed LLMs 
    through the lens of conformal prediction, providing a 
    distribution-free and statistically principled measure of how 
    compression affects model reliability beyond accuracy.
    
    \item \textbf{A unified benchmark across compression paradigms.} 
    We construct a comprehensive benchmark covering 12 LLMs from 
    four families (1B to 70B parameters), various compression 
    configurations spanning weight-only quantization, 
    weight-activation quantization, unstructured pruning, and 
    structured pruning, evaluated on five standard NLP tasks under 
    an identical protocol that isolates the effect of compression.
    
    \item \textbf{Three key findings with practical implications.} 
    Our analysis reveals that (i) compression may decouple accuracy 
    preservation from uncertainty preservation; (ii) larger models 
    tend to absorb compression-induced uncertainty inflation more 
    effectively than smaller ones; and (iii) this inflation is often 
    threshold-like rather than gradual, offering concrete guidance 
    for the safe deployment of compressed LLMs.
 
\end{itemize}

\section{Related Work}
\textbf{Model Compression.}
Quantization~\cite{tong2025robust,tong2026enhancing,zhang2026forget,tong2025data} and pruning~\cite{frantar2023sparsegpt,sun2023simple} are the two dominant paradigms for compressing LLMs. On the quantization side, GPTQ~\cite{frantar2022gptq}  minimizes layer-wise reconstruction error using approximate Hessian information, while AWQ~\cite{lin2024awq}  protects salient weight channels identified through activation magnitudes. More recently, rotation-based methods have emerged to suppress outliers prior to quantization: QuaRot~\cite{ashkboos2024quarot}  applies randomized Hadamard transformations for end-to-end 4-bit inference, SpinQuant~\cite{liu2024spinquant} formulates rotation-based quantization as a constrained optimization problem, learning quantization-friendly orthogonal rotation matrices via Cayley SGD, in line with broader optimization-based approaches~\cite{chang2025convergence,chang2026muoneq,chang2026mgup}. FlatQuant~\cite{sun2025flatquant}  flattens weight and activation distributions through learnable per-layer affine transformations. Beyond rotation-based approaches, SmoothQuant~\cite{xiao2023smoothquant} migrates quantization difficulty from activations to weights via a mathematically equivalent transformation, enabling efficient W8A8 quantization; LLM.int8()~\cite{dettmers2022gpt3} uses mixed-precision decomposition to handle activation outliers; OmniQuant~\cite{shao2024omniquant} learns omnidirectionally calibrated quantization parameters. Quantization has also been integrated with parameter-efficient fine-tuning through QLoRA~\cite{dettmers2023qlora}, and with efficient serving systems such as ATOM~\cite{zhao2024atom}. On the pruning side, SparseGPT~\cite{frantar2023sparsegpt}  and Wanda~\cite{sun2023simple} achieve high unstructured sparsity through one-shot weight removal, while LLM-Pruner~\cite{ma2023llm} and SliceGPT~\cite{ashkboos2024slicegpt}  perform structured pruning by eliminating entire architectural components. These methods have been extensively evaluated on accuracy, but their impact on uncertainty remains largely unexplored.

\textbf{Uncertainty Quantification and Conformal Prediction.}
Existing approaches to LLM uncertainty quantification span several paradigms, including verbalized confidence~\cite{yang2024verbalized}, ensemble-based methods~\cite{polikar2006ensemble,lakshminarayanan2017simple}, Monte Carlo Dropout~\cite{serpell2019probabilistic,gal2016dropout}, token-level entropy~\cite{chen2025ares}, evidential deep learning~\cite{sensoy2018evidential}, and semantic uncertainty~\cite{kuhn2023semantic}. While these methods offer useful heuristics, they lack formal statistical guarantees. Moreover, modern neural networks have been shown to be poorly calibrated~\cite{guo2017calibration,minderer2021revisiting}, further complicating reliable uncertainty estimation. Conformal prediction~\cite{shafer2008tutorial,angelopoulos2023conformal} addresses this limitation by providing distribution-free, finite-sample coverage guarantees for any predictive model. Originally developed for traditional machine learning, conformal prediction has recently been extended to deep learning and, more recently, to LLMs. \cite{ye2024benchmarking} established the first benchmark evaluating LLM uncertainty through conformal prediction, reporting coverage rate and prediction set size across multiple model families and tasks. \cite{wang2025copu} and \cite{kumar2023conformal} further explored conformal methods for text generation and question answering, respectively. \cite{ulmer2024non} addressed the non-exchangeability challenge in autoregressive conformal prediction, and \cite{cherian2024large} proposed enhanced conformal methods for LLM validity. For a comprehensive survey of uncertainty quantification and calibration in LLMs~\cite{liu2025uncertainty}. On the compression side,~\cite{zhong2025quantized} studied calibration degradation under quantization using CCTP and VCAA methods, but their analysis is limited to calibration metrics without conformal prediction guarantees and does not consider pruning. All existing conformal prediction studies for LLMs evaluate only full-precision models. Our work bridges these two lines by providing the first systematic evaluation of conformal prediction under both quantization and pruning.

\section{Background}

\subsection{Revisiting Conformal Prediction}

\textbf{Conformal Prediction.}
Conformal prediction is a distribution-free framework that transforms the outputs of any predictive model into prediction sets with formal coverage guarantees~\cite{shafer2008tutorial,angelopoulos2023conformal}. Given a pre-trained model and a user-specified error rate $\alpha \in (0,1)$, conformal prediction constructs a prediction set $\mathcal{C}(X)$ for each input $X$ such that the true label $Y$ is contained in the set with probability at least $1 - \alpha$:
\begin{equation}
    \mathbb{P}[Y \in \mathcal{C}(X)] \geq 1 - \alpha.
\end{equation}
This guarantee holds under the sole assumption that the calibration and test data are exchangeable, requiring no assumptions about the model architecture or the data distribution.

The standard split conformal prediction procedure operates as follows. Given a calibration set $\mathcal{D}_{\text{cal}} = \{(X_i, Y_i)\}_{i=1}^{n}$ and a nonconformity score function $s(X, Y)$ that measures how poorly a prediction matches the true label, a threshold $\hat{q}$ is computed as the $\lceil (1-\alpha)(n+1) \rceil / n$ quantile of the calibration scores $\{s(X_i, Y_i)\}_{i=1}^{n}$. At test time, the prediction set for a new input $X_{\text{test}}$ is constructed as:
\begin{equation}
    \mathcal{C}(X_{\text{test}}) = \{y \in \mathcal{Y} : s(X_{\text{test}}, y) \leq \hat{q}\},
\end{equation}
where $\mathcal{Y}$ denotes the label space.

\textbf{Conformal Score Functions.}
Following \cite{ye2024benchmarking}, we adopt two widely used conformal score functions in our benchmark.

\textit{LAC (Least Ambiguous set-valued Classifiers)}~\cite{sadinle2019least} defines the nonconformity score as $s_{\text{LAC}}(X, Y) = 1 - \hat{\pi}(Y \mid X)$, where $\hat{\pi}(Y \mid X)$ is the estimated probability of the true label $Y$. The resulting prediction set includes all labels whose predicted probability exceeds $1 - \hat{q}$:
\begin{equation}
    \mathcal{C}_{\text{LAC}}(X) = \{y \in \mathcal{Y} : \hat{\pi}(y \mid X) \geq 1 - \hat{q}\}.
\end{equation}
LAC tends to produce smaller prediction sets but may undercover difficult instances.

\textit{APS (Adaptive Prediction Sets)}~\cite{romano2020classification} constructs prediction sets by accumulating class probabilities in descending order. Let $\pi_{(1)}(X) \geq \pi_{(2)}(X) \geq \cdots$ denote the sorted predicted probabilities and $o(Y)$ the rank of the true label $Y$. The nonconformity score is defined as $s_{\text{APS}}(X, Y) = \sum_{k=1}^{o(Y)} \hat{\pi}(y_{(k)} \mid X)$, and the prediction set is:
\begin{equation}
    \mathcal{C}_{\text{APS}}(X) = \{y_{(1)}, \ldots, y_{(K)}\}, \quad K = \min\left\{k : \sum_{j=1}^{k} \hat{\pi}(y_{(j)} \mid X) \geq \hat{q}\right\}.
\end{equation}
APS provides more adaptive coverage across instances of varying difficulty, but generally yields larger prediction sets. Following \cite{ye2024benchmarking}, we report the average of both score functions throughout our experiments to mitigate the influence of any single scoring rule and ensure a more robust assessment of uncertainty.

\subsection{Revisiting Model Quantization and Pruning}

\textbf{Quantization.}
Model quantization~\cite{tong2025robust,tong2026enhancing} reduces memory footprint and accelerates inference by representing model weights (and optionally activations) with lower-precision numerical formats. Given a full-precision weight tensor $\mathbf{W} \in \mathbb{R}^{m \times n}$, a common uniform quantization scheme maps each element to a discrete set of values representable in $b$ bits:
\begin{equation}
    \hat{w} = s \cdot \text{clamp}\left(\left\lfloor \frac{w}{s} \right\rceil + z, \; 0, \; 2^b - 1\right),
\end{equation}
where $s$ is the scale factor, $z$ is the zero-point, and $\lfloor \cdot \rceil$ denotes rounding to the nearest integer. This formulation captures the basic quantization procedure; in practice, advanced methods further introduce preprocessing steps such as rotation transformations~\cite{ashkboos2024quarot} or channel-aware scaling~\cite{lin2024awq} to suppress outliers and reduce quantization error prior to rounding. We detail the specific quantization methods used in our benchmark in Section~\ref{design}.

\textbf{Pruning.}
Model pruning~\cite{frantar2023sparsegpt,sun2023simple} reduces model size by removing redundant or less important parameters, yielding sparse models with fewer effective computations. Pruning methods can be broadly categorized into two paradigms. \textit{Unstructured pruning} removes individual weights by applying a binary mask $\mathbf{M} \in \{0, 1\}^{m \times n}$ to a weight matrix $\mathbf{W}$:
\begin{equation}
    \hat{\mathbf{W}} = \mathbf{M} \odot \mathbf{W},
\end{equation}
where $\odot$ denotes element-wise multiplication, and the sparsity ratio is defined as $1 - \|\mathbf{M}\|_0 / (m \times n)$. The resulting sparse matrices preserve the original architecture but require specialized hardware or software support for acceleration. \textit{Structured pruning}, by contrast, removes entire architectural components such as attention heads, neurons, or layers, producing smaller but dense models that are directly compatible with standard hardware. The two paradigms differ fundamentally in their compression mechanisms and hardware requirements, making their respective impacts on uncertainty an important axis of comparison. We describe the specific pruning methods used in our benchmark in Section~\ref{design}.

\section{Benchmark Design}
\label{design}

\textbf{Tasks and Datasets.}
To enable a direct and controlled comparison with full-precision baselines, we adopt the same five evaluation tasks and datasets as \cite{ye2024benchmarking}: question answering (MMLU~\cite{hendrycks2020measuring}), reading comprehension (CosmosQA~\cite{huang2019cosmos}), commonsense inference (HellaSwag~\cite{zellers2019hellaswag}), dialogue response selection (HaluDial~\cite{li2023halueval}), and document summarization (HaluSum~\cite{li2023halueval}). Following \cite{ye2024benchmarking}, all tasks are formulated as multiple-choice questions with a standardized set of six options. MMLU, CosmosQA, and HellaSwag originally contain four options per question, while HaluDial and HaluSum contain only two. For HaluDial and HaluSum, two additional choices are first added by randomly sampling from other questions in the same dataset. All five datasets are then augmented with two further options, ``I don't know'' and ``None of the above'', resulting in six options per question. This standardization increases task difficulty and enables finer-grained uncertainty quantification. Due to the computational cost of evaluating numerous compressed model variants, we randomly sample 2,000 instances from each dataset, with 50\% used for calibration and 50\% for testing. By keeping the evaluation protocol identical across all models, any observed differences in accuracy or uncertainty can be attributed solely to the effect of model compression.

\textbf{Evaluation Models.}
We select 12 models spanning four representative LLM families:
the Llama-2 series~\cite{touvron2023llama} with 7B, 13B, and 70B variants,
the Llama-3 series~\cite{grattafiori2024llama} including Llama-3.2-1B, Llama-3.1-8B, and Llama-3.1-70B,
the Qwen-3 series~\cite{yang2025qwen3} including 8B, 14B, 32B, and 30B-A3B variants,
DeepSeek-7B~\cite{bi2024deepseek}, and Falcon-7B~\cite{almazrouei2023falcon}.
The selected models cover a wide range of scales from 1B to 70B parameters and include both dense architectures and the Mixture-of-Experts architecture Qwen3-30B-A3B.

\textbf{Compression Configurations.}
Our benchmark covers two major compression paradigms, each with two representative settings.

\textit{Quantization.}
We consider two quantization settings that differ in whether activations are also quantized.
Weight-only quantization (W4A16) quantizes weights to 4-bit while keeping activations in 16-bit, using three methods: RTN (Round-To-Nearest), AWQ~\cite{lin2024awq}, and GPTQ~\cite{frantar2022gptq}.
Weight-activation quantization (W4A4) quantizes both weights and activations to 4-bit, using three rotation-based methods: QuaRot~\cite{ashkboos2024quarot}, SpinQuant~\cite{liu2024spinquant}, and FlatQuant~\cite{sun2025flatquant}.

\textit{Pruning.}
We consider two pruning paradigms that differ in the granularity of removal.
Unstructured pruning removes individual weights at 50\% sparsity, using three methods: Magnitude pruning, SparseGPT~\cite{frantar2023sparsegpt}, and Wanda~\cite{sun2023simple}.
Structured pruning removes entire architectural components at 20\% sparsity, using two methods: LLM-Pruner~\cite{ma2023llm} and SliceGPT~\cite{ashkboos2024slicegpt}.

We additionally conduct experiments combining quantization and pruning, with results provided in the Appendix.

\textbf{Evaluation Metrics.}
Following~\cite{ye2024benchmarking}, we evaluate all models along three dimensions. \textit{Prediction Accuracy} (Acc) measures the proportion of correctly predicted answers. \textit{Coverage Rate} (CR) verifies whether the conformal prediction sets satisfy the coverage guarantee $\mathrm{CR} \geq 1 - \alpha$, serving as a sanity check for the validity of the prediction sets. \textit{Set Size} (SS) measures the average number of labels in the prediction set and serves as our primary uncertainty metric --- smaller sets indicate higher confidence, while larger sets signal greater uncertainty. We set the error rate $\alpha = 0.1$ throughout all experiments. All reported results are averaged over the LAC and APS score functions as well as the three prompting strategies described in Section~\ref{Per-Score} and~\ref{prompt}.

\textbf{Evaluation Setup.}
We follow the evaluation protocol of \cite{ye2024benchmarking} unless otherwise specified. For each task, we randomly sample 2,000 instances due to the computational cost of evaluating many compressed model variants. Each dataset is split evenly, with 50\% used as the calibration set and the remaining 50\% used as the test set. All tasks are formulated as six-option multiple-choice questions. Specifically, MMLU, CosmosQA, and HellaSwag retain their original four options, while HaluDial and HaluSum are augmented with two randomly sampled distractor options; we further add ``I don't know'' and ``None of the above'' to all datasets. We set the conformal error rate to $\alpha=0.1$, corresponding to a target coverage level of 90\%. We use the same prompting and scoring protocol as \cite{ye2024benchmarking}. For each input, we compute the logits of option letters A--F at the final token position and apply softmax over these logits to obtain option probabilities. Results are averaged over two conformal score functions, LAC and APS, and three prompting strategies: base prompt, shared instruction prompt, and task-specific instruction prompt. Following the original setup, we use five demonstrations for QA, RC, and CI, three demonstrations for DRS, and one demonstration for DS, with a maximum input length of 2048 tokens. We report coverage rate (CR), prediction accuracy (Acc), and prediction set size (SS) on the test set. All experiments are conducted on a server equipped with 8 NVIDIA H100 GPUs.

\section{Experiments and Analysis}

\begin{table*}[t]
\centering
\caption{Task-averaged prediction accuracy (Acc~\%) and prediction set size (SS) across nine models under various compression configurations. All values are averaged over five tasks. ``--'': not evaluated. $\dagger$: degenerate output (CR~$\approx$~12\%). Per-task breakdowns and Qwen3 results are in the Appendix.}
\label{table:overall}
\scriptsize
\setlength{\tabcolsep}{1.8pt}
\begin{tabular}{ll cc cc cc cc cc cc cc cc cc}
\toprule
\multirow{2}{*}{\textbf{Config}} &
\multirow{2}{*}{\textbf{Method}} &
\multicolumn{2}{c}{\textbf{L2-7B}} &
\multicolumn{2}{c}{\textbf{L2-13B}} &
\multicolumn{2}{c}{\textbf{L2-70B}} &
\multicolumn{2}{c}{\textbf{L3-1B}} &
\multicolumn{2}{c}{\textbf{L3-8B}} &
\multicolumn{2}{c}{\textbf{L3-70B}} &
\multicolumn{2}{c}{\textbf{Q3-8B}} &
\multicolumn{2}{c}{\textbf{DS-7B}} &
\multicolumn{2}{c}{\textbf{Fal-7B}} \\
\cmidrule(lr){3-4} \cmidrule(lr){5-6} \cmidrule(lr){7-8} \cmidrule(lr){9-10} \cmidrule(lr){11-12} \cmidrule(lr){13-14} \cmidrule(lr){15-16} \cmidrule(lr){17-18} \cmidrule(lr){19-20}
& & Acc & SS & Acc & SS & Acc & SS & Acc & SS & Acc & SS & Acc & SS & Acc & SS & Acc & SS & Acc & SS \\
\midrule
FP16 & -- & 47.09 & 3.09 & 60.72 & 2.60 & 72.48 & 2.17 & 29.06 & 3.57 & 66.27 & 2.45 & 77.60 & 1.92 & 70.44 & 2.68 & 45.12 & 3.22 & 24.43 & 3.75 \\
\midrule
\multicolumn{20}{c}{\textit{Weight-Only Quantization (W4A16)}} \\
\midrule
W4A16 & RTN       & 44.81 & 3.13 & 59.45 & 2.79 & 71.49 & 2.23 & 27.09 & 3.60 & 66.38 & 2.46 & 71.69 & 2.14 & 67.18 & 2.92 & -- & -- & -- & -- \\
W4A16 & AWQ       & 43.18 & 3.19 & 59.73 & 2.80 & 73.20 & 2.18 & 25.92 & 3.76 & 67.60 & 2.53 & 75.11 & 2.14 & 69.61 & 2.68 & 43.07 & 3.29 & 22.63 & 3.89 \\
W4A16 & GPTQ      & 46.20 & 3.11 & 58.79 & 2.69 & 69.11 & 2.22 & 25.82 & 3.61 & 66.35 & 2.35 & 77.82 & 1.96 & 68.33 & 2.83 & -- & -- & -- & -- \\
\midrule
\multicolumn{20}{c}{\textit{Weight-Activation Quantization (W4A4)}} \\
\midrule
W4A4  & FlatQuant & 28.63 & 3.75 & 46.49 & 3.08 & 62.40 & 2.48 & 24.09 & 4.40 & 36.05 & 3.70 & -- & -- & 17.61 & 5.17 & -- & -- & -- & -- \\
W4A4  & QuaRot    & 22.04 & 5.65 & 45.90 & 3.32 & 66.49 & 2.54 & 20.42 & 5.51 & 45.79 & 3.20 & -- & -- & 18.26 & 5.46 & 35.06 & 3.72 & -- & -- \\
W4A4  & SpinQuant & 26.26 & 4.52 & 47.56 & 3.08 & 69.01 & 2.43 & 24.78 & 3.98 & 49.63 & 3.02 & -- & -- & -- & -- & -- & -- & -- & -- \\
\midrule
\multicolumn{20}{c}{\textit{Unstructured Pruning (50\%)}} \\
\midrule
     FP16  & Magnitude  & 24.98 & 4.55 & 43.95 & 3.10 & 60.64 & 2.73 & 24.51 & 4.25 & 25.83 & 4.18 & 55.84 & 2.98 & 29.79 & 3.54 & 24.58 & 3.64 & 25.21 & 5.02 \\
     FP16  & SparseGPT  & 34.13 & 3.48 & 46.12 & 3.04 & 65.30 & 2.44 & 25.26 & 3.58 & 47.27 & 3.01 & 65.20 & 2.33 & 64.09 & 2.73 & 34.15 & 3.40 & 24.47 & 3.75 \\
     FP16  & Wanda      & 31.13 & 3.58 & 49.11 & 2.97 & 66.63 & 2.42 & 25.28 & 3.65 & 43.70 & 2.94 & 64.90 & 2.29 & 63.35 & 2.83 & 34.77 & 3.40 & 24.55 & 3.65 \\
\midrule
\multicolumn{20}{c}{\textit{Structured Pruning (20\%)}} \\
\midrule
     FP16  & LLM-Pruner & 25.22 & 5.74 & 31.98 & 4.07 & 67.39 & 2.60 & 24.13 & 4.59 & 48.36 & 3.04 & 74.25 & 2.13 & 64.54 & 2.73 & 18.45 & 5.40 & 23.01 & 3.97 \\
     FP16  & SliceGPT   & 26.43 & 3.74 & 39.02 & 3.65 & 57.33 & 2.70 & 24.87 & 3.96 & 36.56 & 3.53 & 58.51 & 2.62 & 25.22$^\dagger$ & 1.00$^\dagger$ & 33.56 & 3.47 & 25.07 & 4.66 \\
\bottomrule
\end{tabular}
\end{table*}

\subsection{Overall Results}

Table~\ref{table:overall} summarizes the task-averaged accuracy (Acc) and prediction set size (SS) across all evaluated models and compression settings. Overall, weight-only quantization is the most stable regime: W4A16 usually keeps both Acc and SS close to the FP16 baseline, especially for medium and large models. For example, Llama2-70B under AWQ achieves 73.20 Acc and 2.18 SS, nearly matching the FP16 baseline of 72.48 Acc and 2.17 SS, while Qwen3-8B under AWQ remains close to its dense baseline (69.61 Acc and 2.68 SS versus 70.44 Acc and 2.68 SS). In contrast, W4A4 is substantially more disruptive, producing larger accuracy drops and higher uncertainty. On Llama2-7B, QuaRot reduces Acc from 47.09 to 22.04 while increasing SS from 3.09 to 5.65, whereas Llama2-70B under the same setting retains 66.49 Acc with SS of 2.54, indicating that larger models better absorb aggressive activation quantization.

Pruning exhibits a more heterogeneous pattern. Under 50\% unstructured pruning, Magnitude pruning is consistently the weakest method, while SparseGPT and Wanda better preserve Acc and SS but remain sensitive to model scale; for instance, Wanda yields 66.63 Acc and 2.42 SS on Llama2-70B but drops to 31.13 Acc and 3.58 SS on Llama2-7B. Structured pruning at 20\% also varies across methods and architectures: LLM-Pruner remains relatively effective on larger models such as Llama3.1-70B, whereas SliceGPT is unstable on Qwen3-8B, where the degenerate result marked by $\dagger$ indicates unreliable uncertainty. Taken together, Table~\ref{table:overall} shows that compression quality cannot be judged by accuracy alone: W4A16 is generally the safest setting for uncertainty preservation, W4A4 causes the largest uncertainty inflation, and pruning requires careful method selection.

\subsection{Accuracy–Uncertainty Decoupling}
\label{sec:decoupling}

\vspace{0.2cm}
\begin{mdframed}[
  backgroundcolor=KFbg,
  linecolor=KFblue,
  linewidth=0pt,
  leftline=true,
  linewidth=3pt,
  innerleftmargin=3mm,
  innerrightmargin=3mm,
  innertopmargin=2mm,
  innerbottommargin=2mm,
  skipabove=0.25cm,
  skipbelow=0.25cm
]
\textbf{\textcolor{KFblue}{Key Finding 1: }}
\textit{Compression tends to  decouple 
accuracy from uncertainty.}
\end{mdframed}
\vspace{0.1cm}

\begin{figure*}[t]
\centering
\includegraphics[width=\textwidth]{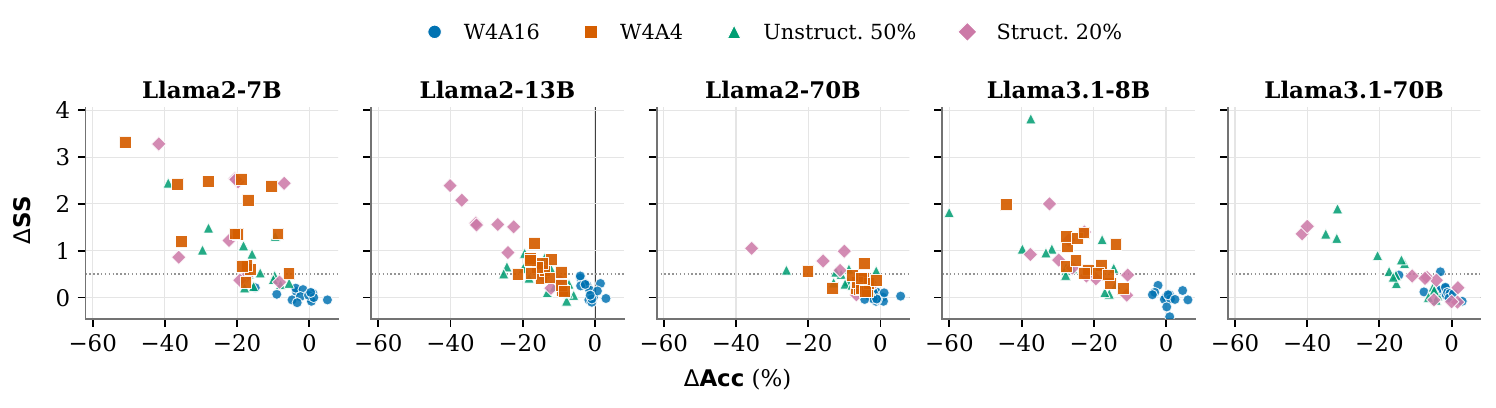}
\caption{Accuracy change ($\Delta$Acc) vs.\ uncertainty change ($\Delta$SS) relative to uncompressed baselines for five models across all compression methods and tasks. If accuracy and uncertainty were coupled, points would concentrate along a consistent negative-slope trend; instead, substantial dispersion is observed in all subplots, confirming that compression frequently decouples the two metrics (Finding~1). Comparing across subplots, the scatter contracts toward the origin as model scale increases, further illustrating that larger models absorb compression-induced uncertainty more effectively (Finding~2).}
\label{fig:acc-ss-scatter}
\end{figure*}

A natural assumption is that compressed models lose accuracy and uncertainty quality in tandem: if the top-1 prediction remains reliable, the predictive distribution should also remain well calibrated. Figure~\ref{fig:acc-ss-scatter} shows that this assumption does not generally hold. We compute $\Delta\mathrm{Acc}=\mathrm{Acc}_{\mathrm{compressed}}-\mathrm{Acc}_{\mathrm{dense}}$ and $\Delta\mathrm{SS}=\mathrm{SS}_{\mathrm{compressed}}-\mathrm{SS}_{\mathrm{dense}}$ for each model, compression method, and task in Tables~\ref{tab:llama2-quant}--\ref{tab:compare-main-other-models-combined}. Rather than forming a compact negative-slope trend, the points are widely dispersed, indicating that compression often changes prediction accuracy and uncertainty in different ways.

This decoupling appears in two forms. First, accuracy may remain close to the dense baseline while SS increases: for Llama2-13B on Commonsense Inference, W4A16 AWQ slightly improves accuracy from $59.63\%$ to $61.17\%$, yet SS increases from $2.83$ to $3.13$; W4A16 RTN similarly changes accuracy from $59.63\%$ to $56.77\%$ while increasing SS from $2.83$ to $3.13$. Second, accuracy may drop sharply while SS changes little: for Llama3.1-8B under $50\%$ Wanda pruning, QA accuracy drops from $62.93\%$ to $45.97\%$, whereas SS changes only from $2.98$ to $3.09$. Overall, W4A16 points are concentrated near the origin, W4A4 points move farther away especially for smaller models, and pruning exhibits heterogeneous behavior, showing that accuracy-only evaluation is insufficient for uncertainty-sensitive deployment.

\subsection{The Role of Model Scale}
\label{sec:scale}

\vspace{0.2cm}
\begin{mdframed}[
  backgroundcolor=KFbg,
  linecolor=KFblue,
  linewidth=0pt,
  leftline=true,
  linewidth=3pt,
  innerleftmargin=3mm,
  innerrightmargin=3mm,
  innertopmargin=2mm,
  innerbottommargin=2mm,
  skipabove=0.25cm,
  skipbelow=0.25cm
]
\textbf{\textcolor{KFblue}{Key Finding 2: }}
\textit{Larger models tend to absorb compression-induced uncertainty better.}
\end{mdframed}
\vspace{0.1cm}

Model scale strongly affects how much uncertainty inflation compression induces. Under W4A4 QuaRot quantization in Table~\ref{tab:llama2-quant}, the average SS increase across five tasks is approximately $+2.56$ for Llama2-7B, $+0.72$ for Llama2-13B, and $+0.37$ for Llama2-70B, giving about a $6.9\times$ reduction in uncertainty inflation from 7B to 70B. The same trend is visible at the task level: on QA, SS changes from $3.20$ to $5.73$ for Llama2-7B, from $3.10$ to $3.36$ for Llama2-13B, and from $2.64$ to $2.80$ for Llama2-70B.

The scale effect also appears under pruning. With $50\%$ Wanda pruning on RC, Table~\ref{tab:llama2-pruning} shows SS increases of $+1.02$ for Llama2-7B ($2.46\rightarrow3.48$), $+0.23$ for Llama2-13B ($2.33\rightarrow2.56$), and $+0.15$ for Llama2-70B ($1.79\rightarrow1.94$). In the Llama3 family, Llama3.1-8B RC SS increases from $1.89$ to $2.53$, while Llama3.1-70B RC SS slightly decreases from $1.61$ to $1.56$. However, scale is not a complete safeguard: Llama3.1-70B under Magnitude pruning still shows a large CI SS increase from $1.89$ to $3.25$, suggesting that larger models provide redundancy but can still fail on task-specific uncertainty-critical behavior.

\begin{figure*}[t]
\centering
\includegraphics[width=\textwidth]{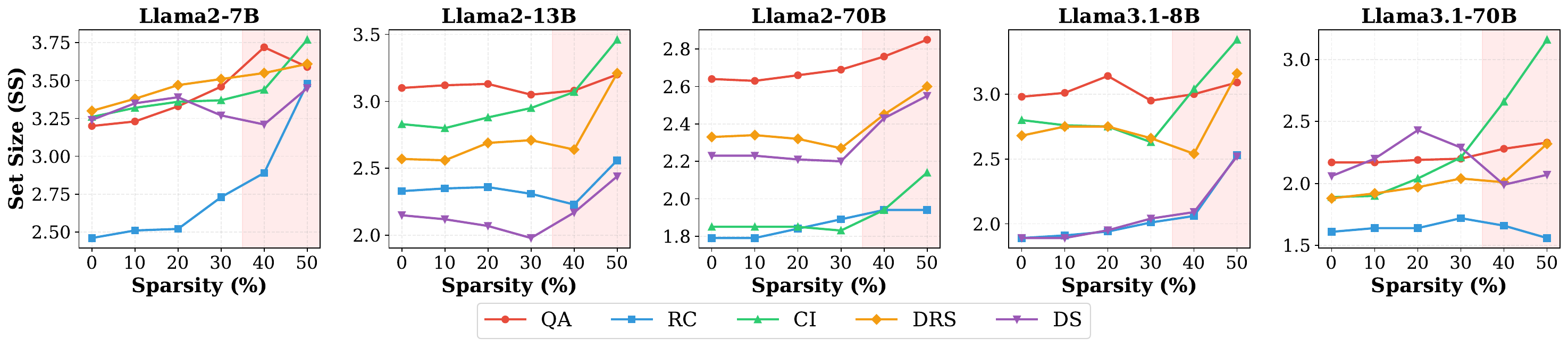}
\caption{Prediction set size (SS) as a function of Wanda pruning sparsity (0\%--50\%) across five tasks for five models spanning two families and three scales. The shaded region marks the high-sparsity regime (40\%--50\%) where SS often inflates sharply rather than gradually, illustrating the threshold-like behavior described in Finding~3. Comparing across subplots further reveals the scale effect (Finding~2): 70B models maintain consistently lower and flatter SS trajectories than their smaller counterparts under identical compression.}
\label{fig:ss-vs-sparsity}
\end{figure*}

\subsection{Threshold-Like Uncertainty Inflation}
\label{sec:threshold}

\vspace{0.2cm}
\begin{mdframed}[
  backgroundcolor=KFbg,
  linecolor=KFblue,
  linewidth=0pt,
  leftline=true,
  linewidth=3pt,
  innerleftmargin=3mm,
  innerrightmargin=3mm,
  innertopmargin=2mm,
  innerbottommargin=2mm,
  skipabove=0.25cm,
  skipbelow=0.25cm
]
\textbf{\textcolor{KFblue}{Key Finding 3: }}
\textit{Uncertainty inflation is often threshold-like rather than gradual.}
\end{mdframed}
\vspace{0.1cm}

To examine how uncertainty changes with compression intensity, Figure~\ref{fig:ss-vs-sparsity} plots SS under Wanda pruning from $0\%$ to $50\%$ sparsity, with numerical values reported in Tables~\ref{tab:llama2-progressive} and~\ref{tab:llama3-progressive}. The trajectories are often not linear. For Llama3.1-8B, RC SS remains within $1.89$--$2.06$ from $0\%$ to $40\%$, then jumps to $2.53$ at $50\%$; CI remains near $2.75$--$2.80$ through $20\%$, then rises to $3.04$ and $3.42$ at $40\%$ and $50\%$; DRS decreases to $2.54$ at $40\%$ before jumping to $3.16$ at $50\%$. Llama2-7B shows a similar RC transition, where the $40\%$ to $50\%$ step alone increases SS by $+0.59$ ($2.89\rightarrow3.48$), accounting for more than half of the total $0\%$--$50\%$ inflation.

The threshold location depends on both model scale and task. Larger models are flatter on many tasks: Llama2-70B RC changes only from $1.79$ to $1.94$ across the full sparsity range. Yet even 70B models are not immune, as Llama3.1-70B CI rises from $1.89$ at $0\%$ to $2.21$ at $30\%$, then jumps to $2.66$ at $40\%$ and $3.16$ at $50\%$. These results suggest that moderate pruning can remove redundant parameters with limited uncertainty impact, but once a task-specific tipping point is crossed, SS can inflate abruptly. Therefore, compressed models should be evaluated across a sweep of sparsity levels, since a configuration that appears stable at $30\%$ may become unreliable at $40\%$ or $50\%$.

\subsection{MoE Results}
\label{sec:moe-results}

Qwen3-30B-A3B provides a representative MoE case, where only a subset of experts is activated for each input and compression may affect both expert-specific capacity and routing-dependent behavior. As shown in Tables~\ref{tab:compare-main-qwen-moe} and~\ref{tab:compare-main-qwen3-moe-combined}, the dense MoE baseline obtains an average Acc of $46.52$ and an average SS of $3.32$ across five tasks. Among W4A16 methods, RTN is the most stable, with average Acc changing only from $46.52$ to $45.64$ and average SS decreasing from $3.32$ to $3.09$, whereas AWQ reduces average Acc to $37.17$ and increases SS to $3.59$. GPTQ appears strong on average, with Acc $48.40$ and SS $2.75$, but this should be interpreted carefully because the DS entry has very low coverage ($12.68$) and a degenerate SS of $1.00$. Pruning results are more polarized: under $50\%$ unstructured pruning, Magnitude and Wanda maintain or improve average Acc ($47.70$ and $50.07$) but increase SS to $3.81$ and $3.64$, while SparseGPT lowers Acc to $34.07$ with SS close to the dense baseline ($3.39$). Structured pruning shows the strongest contrast, with LLM-Pruner at $20\%$ improving both Acc and SS ($74.33$ Acc, $2.01$ SS), whereas SliceGPT collapses to $16.05$ Acc and inflates SS to $5.57$. These results suggest that MoE compression is highly method-dependent, and that sparsity or bit-width alone is insufficient to predict uncertainty reliability.

\subsection{Ablation Results for Llama2-7B with Different Precision}
\label{sec:ablation-llama2-7b}

Table~\ref{tab:ablation-llama2-7b} isolates the effect of activation--weight precision on Llama2-7B. The FP16 baseline has average Acc $47.09$ and SS $3.09$. Moderate quantization remains relatively stable: W8A8 QuaRot slightly improves average Acc to $47.20$ but increases SS to $3.33$, W8A8 SmoothQuant reduces Acc mildly to $45.50$ and keeps SS close to baseline at $3.16$, and W6A6 QuaRot gives Acc $45.36$ with SS $3.35$. In contrast, W4A4 SpinQuant causes a sharp degradation, reducing average Acc to $26.26$ and increasing average SS to $4.52$; the RC task is especially affected, with Acc dropping from $67.20$ to $30.72$ and SS rising from $2.46$ to $4.88$. This ablation supports the broader finding that uncertainty degradation is not gradual with precision: Llama2-7B tolerates W8A8 and W6A6 relatively well, but W4A4 crosses into a much more damaging regime for both accuracy and uncertainty.

\begin{table*}[hptb]
\centering
\caption{Additional Ablation Results for Llama2-7B with Different Precision}
\scriptsize
\label{tab:ablation-llama2-7b}
\setlength{\tabcolsep}{2.4pt}
\begin{tabular}{llc ccc ccc ccc ccc ccc}
\toprule \multirow{2}{*}{\textbf{Model}}
 &
  \multirow{2}{*}{\textbf{Bit}} &
  \multirow{2}{*}{\textbf{Method}} &
  \multicolumn{3}{c}{\textbf{QA}} &
  \multicolumn{3}{c}{\textbf{RC}} &
  \multicolumn{3}{c}{\textbf{CI}} &
  \multicolumn{3}{c}{\textbf{DRS}} &
  \multicolumn{3}{c}{\textbf{DS}} \\
  \cmidrule(lr){4-6} \cmidrule(lr){7-9} \cmidrule(lr){10-12} \cmidrule(lr){13-15} \cmidrule(lr){16-18}
  & & &
  CR & Acc & SS &
  CR & Acc & SS &
  CR & Acc & SS &
  CR & Acc & SS &
  CR & Acc & SS \\
\midrule
\multirow{5}{*}{Llama2-7B}
  & FP16   & --        & 91.40 & 45.53 & 3.20 & 91.83 & 67.20 & 2.46 & 90.47 & 43.27 & 3.26 & 90.73 & 32.67 & 3.30 & 89.85 & 46.77 & 3.24\\
  \noalign{\vspace{0.1em}}\cdashline{2-18}\noalign{\vspace{0.1em}}
  & W8A8 & QuaRot       & 91.90 & 44.37 & 3.20 & 90.95 & 65.07 & 2.69 & 90.23 & 41.93 & 3.37 & 90.37 & 32.97 & 3.82 & 92.37 & 51.67 & 3.57\\
  & W8A8 & SmoothQuant       &91.45 & 44.97 & 3.20 & 91.98 & 66.57 & 2.46 & 90.70 & 39.37 & 3.32 & 89.96 & 33.30 & 3.27 & 92.65 & 43.29 & 3.56 \\
  & W6A6  & QuaRot       &91.50 & 43.70 & 3.22 & 91.32 & 62.73 & 2.82 & 90.00 & 39.83 & 3.40 & 90.52 & 33.40 & 3.76 & 91.83 & 47.16 & 3.55 \\
  & W4A4  & SpinQuant      &89.36&25.77&4.55&90.39&30.72&4.88&89.86&24.70&3.93&88.86&23.96&4.66&90.39&26.17&4.60 \\
\midrule
\end{tabular}%
\end{table*}

\subsection{Discussion}
\paragraph{Quantization vs. pruning.}
W4A16 quantization is the most uncertainty-preserving compression regime: across model scales and tasks, it usually keeps SS close to the FP16 baseline and rarely increases it by more than 0.5. In contrast, W4A4 quantization often causes much larger uncertainty inflation, especially for smaller models, where QuaRot and SpinQuant can push SS above 5.0. Pruning shows stronger method dependence. Advanced unstructured methods such as SparseGPT and Wanda are generally more stable than Magnitude pruning at the same 50\% sparsity, while structured pruning is more volatile: LLM-Pruner can preserve SS on larger models, whereas SliceGPT may cause severe distributional collapse. Overall, uncertainty-sensitive deployment should not choose between quantization and pruning based on accuracy alone; W4A16 is the safest default, while pruning requires method-specific uncertainty validation.
\paragraph{Practical recommendations.} Based on our findings, we distill several guidelines for deploying compressed LLMs in uncertainty-sensitive settings. First, accuracy preservation should not be treated as a proxy for uncertainty preservation---explicit SS evaluation under conformal prediction should be a standard step in compression pipelines (\S\ref{sec:decoupling}). Second, when aggressive compression is required, prioritizing larger base models yields substantially better uncertainty robustness, even if a smaller model meets the accuracy target (\S\ref{sec:scale}). Third, compressed models should be evaluated across a sweep of compression ratios rather than at a single operating point, since the threshold-like nature of uncertainty inflation means that a seemingly safe configuration may be close to a tipping point (\S\ref{sec:threshold}). More broadly, these guidelines align with the growing call for holistic, multi-dimensional evaluation of LLMs~\cite{liang2022holistic,chang2024survey}, and we advocate that uncertainty should be a standard axis alongside accuracy, efficiency, and fairness.
\paragraph{Limitations.} Our benchmark has several limitations that suggest directions for future work. First, all tasks are formulated as multiple-choice classification, which enables conformal prediction but does not capture open-ended text generation, where uncertainty manifests differently (e.g., through sequence-level entropy or semantic equivalence classes~\cite{kuhn2023semantic}); extending conformal methods to free-form generation~\cite{quach2024conformal} under compression is an important direction for future work. Second, we fix the error rate at $\alpha = 0.1$ and the calibration--test split at 50/50; varying these parameters may reveal additional sensitivity patterns.

\section{Conclusion}
In this paper, we present the first comprehensive benchmark 
for evaluating the uncertainty of compressed LLMs through the 
lens of conformal prediction. As compressed models are 
increasingly deployed in safety-critical settings such as 
medical diagnosis, legal reasoning, and autonomous 
decision-making, evaluating them solely by accuracy risks 
overlooking silent degradations in reliability that only 
surface as inflated uncertainty. Our analysis reveals that 
compression may decouple accuracy from uncertainty, that 
model scale tends to buffer compression-induced uncertainty 
inflation, and that this inflation is often threshold-like 
rather than gradual. Together, these findings establish 
uncertainty as an indispensable axis for evaluating 
compressed LLMs and motivate a rethinking of how compression 
methods are designed and selected. We hope this benchmark 
serves as a foundation for future work on uncertainty-aware 
compression, and on extending conformal prediction to broader 
compression paradigms and generative tasks.

\clearpage

\bibliographystyle{IEEEtran}  % or the downloaded .bst file name
\bibliography{main}

@article{ye2024benchmarking,
  title={Benchmarking llms via uncertainty quantification},
  author={Ye, Fanghua and Yang, Mingming and Pang, Jianhui and Wang, Longyue and Wong, Derek F and Yilmaz, Emine and Shi, Shuming and Tu, Zhaopeng},
  journal={Advances in Neural Information Processing Systems},
  volume={37},
  pages={15356--15385},
  year={2024}
}

@article{grattafiori2024llama,
  title={The {Llama} 3 herd of models},
  author={Grattafiori, Aaron and Dubey, Abhimanyu and Jauhri, Abhinav and Pandey, Abhinav and Kadian, Abhishek and Al-Dahle, Ahmad and Letman, Aiesha and Mathur, Akhil and Schelten, Alan and Vaughan, Alex and others},
  journal={arXiv preprint arXiv:2407.21783},
  year={2024}
}

@article{achiam2023gpt,
  title={Gpt-4 technical report},
  author={Achiam, Josh and Adler, Steven and Agarwal, Sandhini and Ahmad, Lama and Akkaya, Ilge and Aleman, Florencia Leoni and Almeida, Diogo and Altenschmidt, Janko and Altman, Sam and Anadkat, Shyamal and others},
  journal={arXiv preprint arXiv:2303.08774},
  year={2023}
}

@article{sadinle2019least,
  title={Least ambiguous set-valued classifiers with bounded error levels},
  author={Sadinle, Mauricio and Lei, Jing and Wasserman, Larry},
  journal={Journal of the American Statistical Association},
  volume={114},
  number={525},
  pages={223--234},
  year={2019},
  publisher={Taylor \& Francis}
}

@article{romano2020classification,
  title={Classification with valid and adaptive coverage},
  author={Romano, Yaniv and Sesia, Matteo and Candes, Emmanuel},
  journal={Advances in neural information processing systems},
  volume={33},
  pages={3581--3591},
  year={2020}
}

@article{frantar2022gptq,
  title={Gptq: Accurate post-training quantization for generative pre-trained transformers},
  author={Frantar, Elias and Ashkboos, Saleh and Hoefler, Torsten and Alistarh, Dan},
  journal={arXiv preprint arXiv:2210.17323},
  year={2022}
}

@article{lin2024awq,
  title={Awq: Activation-aware weight quantization for on-device llm compression and acceleration},
  author={Lin, Ji and Tang, Jiaming and Tang, Haotian and Yang, Shang and Chen, Wei-Ming and Wang, Wei-Chen and Xiao, Guangxuan and Dang, Xingyu and Gan, Chuang and Han, Song},
  journal={Proceedings of machine learning and systems},
  volume={6},
  pages={87--100},
  year={2024}
}

@article{ashkboos2024quarot,
  title={Quarot: Outlier-free 4-bit inference in rotated llms},
  author={Ashkboos, Saleh and Mohtashami, Amirkeivan and Croci, Maximilian L and Li, Bo and Cameron, Pashmina and Jaggi, Martin and Alistarh, Dan and Hoefler, Torsten and Hensman, James},
  journal={Advances in Neural Information Processing Systems},
  volume={37},
  pages={100213--100240},
  year={2024}
}

@article{liu2024spinquant,
  title={Spinquant: Llm quantization with learned rotations},
  author={Liu, Zechun and Zhao, Changsheng and Fedorov, Igor and Soran, Bilge and Choudhary, Dhruv and Krishnamoorthi, Raghuraman and Chandra, Vikas and Tian, Yuandong and Blankevoort, Tijmen},
  journal={arXiv preprint arXiv:2405.16406},
  year={2024}
}

@inproceedings{sun2025flatquant,
  title={FlatQuant: Flatness Matters for LLM Quantization},
  author={Sun, Yuxuan and Liu, Ruikang and Bai, Haoli and Bao, Han and Zhao, Kang and Li, Yuening and Hu, Jiaxin and Yu, Xianzhi and Hou, Lu and Yuan, Chun and others},
  booktitle={International Conference on Machine Learning},
  pages={57587--57613},
  year={2025},
  organization={PMLR}
}

@inproceedings{frantar2023sparsegpt,
  title={Sparsegpt: Massive language models can be accurately pruned in one-shot},
  author={Frantar, Elias and Alistarh, Dan},
  booktitle={International conference on machine learning},
  pages={10323--10337},
  year={2023},
  organization={PMLR}
}

@article{sun2023simple,
  title={A simple and effective pruning approach for large language models},
  author={Sun, Mingjie and Liu, Zhuang and Bair, Anna and Kolter, J Zico},
  journal={arXiv preprint arXiv:2306.11695},
  year={2023}
}

@article{ma2023llm,
  title={Llm-pruner: On the structural pruning of large language models},
  author={Ma, Xinyin and Fang, Gongfan and Wang, Xinchao},
  journal={Advances in neural information processing systems},
  volume={36},
  pages={21702--21720},
  year={2023}
}

@article{ashkboos2024slicegpt,
  title={Slicegpt: Compress large language models by deleting rows and columns},
  author={Ashkboos, Saleh and Croci, Maximilian L and Nascimento, Marcelo Gennari do and Hoefler, Torsten and Hensman, James},
  journal={arXiv preprint arXiv:2401.15024},
  year={2024}
}

@article{yang2024verbalized,
  title={On verbalized confidence scores for llms},
  author={Yang, Daniel and Tsai, Yao-Hung Hubert and Yamada, Makoto},
  journal={arXiv preprint arXiv:2412.14737},
  year={2024}
}

@article{polikar2006ensemble,
  title={Ensemble based systems in decision making},
  author={Polikar, Robi},
  journal={IEEE Circuits and systems magazine},
  volume={6},
  number={3},
  pages={21--45},
  year={2006},
  publisher={IEEE}
}

@inproceedings{serpell2019probabilistic,
  title={Probabilistic forecasting using Monte Carlo dropout neural networks},
  author={Serpell, Cristi{\'a}n and Araya, Ignacio and Valle, Carlos and Allende, H{\'e}ctor},
  booktitle={Iberoamerican congress on pattern recognition},
  pages={387--397},
  year={2019},
  organization={Springer}
}

@article{chen2025ares,
  title={Ares: Multimodal adaptive reasoning via difficulty-aware token-level entropy shaping},
  author={Chen, Shuang and Guo, Yue and Ye, Yimeng and Huang, Shijue and Hu, Wenbo and Li, Haoxi and Zhang, Manyuan and Chen, Jiayu and Guo, Song and Peng, Nanyun},
  journal={arXiv preprint arXiv:2510.08457},
  year={2025}
}

@article{shafer2008tutorial,
  title={A tutorial on conformal prediction.},
  author={Shafer, Glenn and Vovk, Vladimir},
  journal={Journal of machine learning research},
  volume={9},
  number={3},
  year={2008}
}

@article{angelopoulos2023conformal,
  title={Conformal prediction: A gentle introduction},
  author={Angelopoulos, Anastasios N and Bates, Stephen},
  journal={Foundations and Trends in Machine Learning},
  volume={16},
  number={4},
  pages={494--591},
  year={2023},
  publisher={Emerald Publishing Limited}
}

@inproceedings{zhong2025quantized,
  title={Quantized can still be calibrated: A unified framework to calibration in quantized large language models},
  author={Zhong, Mingyu and Wang, Guanchu and Chuang, Yu-Neng and Zou, Na},
  booktitle={Proceedings of the 63rd Annual Meeting of the Association for Computational Linguistics (Volume 1: Long Papers)},
  pages={30503--30517},
  year={2025}
}

@article{wang2025copu,
  title={Copu: Conformal prediction for uncertainty quantification in natural language generation},
  author={Wang, Sean and Jiang, Yicheng and Tang, Yuxin and Cheng, Lu and Chen, Hanjie},
  journal={arXiv preprint arXiv:2502.12601},
  year={2025}
}

@article{kumar2023conformal,
  title={Conformal prediction with large language models for multi-choice question answering},
  author={Kumar, Bhawesh and Lu, Charlie and Gupta, Gauri and Palepu, Anil and Bellamy, David and Raskar, Ramesh and Beam, Andrew},
  journal={arXiv preprint arXiv:2305.18404},
  year={2023}
}

@inproceedings{tong2025robust,
  title={Robust machine unlearning for quantized neural networks via adaptive gradient reweighting with similar labels},
  author={Tong, Yujia and Wang, Yuze and Yuan, Jingling and Hu, Chuang},
  booktitle={Proceedings of the IEEE/CVF International Conference on Computer Vision},
  pages={20603--20612},
  year={2025}
}

@article{tong2026enhancing,
  title={Enhancing quantization-aware training on edge devices via relative entropy coreset selection and cascaded layer correction},
  author={Tong, Yujia and Yuan, Jingling and Hu, Chuang},
  journal={IEEE Transactions on Mobile Computing},
  year={2026},
  publisher={IEEE}
}

@article{hendrycks2020measuring,
  title={Measuring massive multitask language understanding},
  author={Hendrycks, Dan and Burns, Collin and Basart, Steven and Zou, Andy and Mazeika, Mantas and Song, Dawn and Steinhardt, Jacob},
  journal={arXiv preprint arXiv:2009.03300},
  year={2020}
}

@inproceedings{huang2019cosmos,
  title={Cosmos QA: Machine reading comprehension with contextual commonsense reasoning},
  author={Huang, Lifu and Le Bras, Ronan and Bhagavatula, Chandra and Choi, Yejin},
  booktitle={Proceedings of the 2019 conference on empirical methods in natural language processing and the 9th international joint conference on natural language processing (EMNLP-IJCNLP)},
  pages={2391--2401},
  year={2019}
}

@inproceedings{zellers2019hellaswag,
  title={Hellaswag: Can a machine really finish your sentence?},
  author={Zellers, Rowan and Holtzman, Ari and Bisk, Yonatan and Farhadi, Ali and Choi, Yejin},
  booktitle={Proceedings of the 57th annual meeting of the association for computational linguistics},
  pages={4791--4800},
  year={2019}
}

@inproceedings{li2023halueval,
  title={Halueval: A large-scale hallucination evaluation benchmark for large language models},
  author={Li, Junyi and Cheng, Xiaoxue and Zhao, Xin and Nie, Jian-Yun and Wen, Ji-Rong},
  booktitle={Proceedings of the 2023 conference on empirical methods in natural language processing},
  pages={6449--6464},
  year={2023}
}

@article{touvron2023llama,
  title={Llama 2: Open foundation and fine-tuned chat models},
  author={Touvron, Hugo and Martin, Louis and Stone, Kevin and Albert, Peter and Almahairi, Amjad and Babaei, Yasmine and Bashlykov, Nikolay and Batra, Soumya and Bhargava, Prajjwal and Bhosale, Shruti and others},
  journal={arXiv preprint arXiv:2307.09288},
  year={2023}
}

@article{yang2025qwen3,
  title={Qwen3 technical report},
  author={Yang, An and Li, Anfeng and Yang, Baosong and Zhang, Beichen and Hui, Binyuan and Zheng, Bo and Yu, Bowen and Gao, Chang and Huang, Chengen and Lv, Chenxu and others},
  journal={arXiv preprint arXiv:2505.09388},
  year={2025}
}

@article{bi2024deepseek,
  title={Deepseek llm: Scaling open-source language models with longtermism},
  author={Bi, Xiao and Chen, Deli and Chen, Guanting and Chen, Shanhuang and Dai, Damai and Deng, Chengqi and Ding, Honghui and Dong, Kai and Du, Qiushi and Fu, Zhe and others},
  journal={arXiv preprint arXiv:2401.02954},
  year={2024}
}

@article{almazrouei2023falcon,
  title={The falcon series of open language models},
  author={Almazrouei, Ebtesam and Alobeidli, Hamza and Alshamsi, Abdulaziz and Cappelli, Alessandro and Cojocaru, Ruxandra and Debbah, M{\'e}rouane and Goffinet, {\'E}tienne and Hesslow, Daniel and Launay, Julien and Malartic, Quentin and others},
  journal={arXiv preprint arXiv:2311.16867},
  year={2023}
}

@inproceedings{xiao2023smoothquant,
  title={Smoothquant: Accurate and efficient post-training quantization for large language models},
  author={Xiao, Guangxuan and Lin, Ji and Seznec, Mickael and Wu, Hao and Demouth, Julien and Han, Song},
  booktitle={International conference on machine learning},
  pages={38087--38099},
  year={2023},
  organization={PMLR}
}

@article{dettmers2022gpt3,
  title={Gpt3. int8 (): 8-bit matrix multiplication for transformers at scale},
  author={Dettmers, Tim and Lewis, Mike and Belkada, Younes and Zettlemoyer, Luke},
  journal={Advances in neural information processing systems},
  volume={35},
  pages={30318--30332},
  year={2022}
}

@article{dettmers2023qlora,
  title={Qlora: Efficient finetuning of quantized llms},
  author={Dettmers, Tim and Pagnoni, Artidoro and Holtzman, Ari and Zettlemoyer, Luke},
  journal={Advances in neural information processing systems},
  volume={36},
  pages={10088--10115},
  year={2023}
}

@inproceedings{shao2024omniquant,
  title={Omniquant: Omnidirectionally calibrated quantization for large language models},
  author={Shao, Wenqi and Chen, Mengzhao and Zhang, Zhaoyang and Xu, Peng and Zhao, Lirui and Li, Zhiqian and Zhang, Kaipeng and Peng, Gao and Qiao, Yu and Luo, Ping},
  booktitle={International Conference on Learning Representations},
  volume={2024},
  pages={45472--45496},
  year={2024}
}

@article{zhao2024atom,
  title={Atom: Low-bit quantization for efficient and accurate llm serving},
  author={Zhao, Yilong and Lin, Chien-Yu and Zhu, Kan and Ye, Zihao and Chen, Lequn and Zheng, Size and Ceze, Luis and Krishnamurthy, Arvind and Chen, Tianqi and Kasikci, Baris},
  journal={Proceedings of Machine Learning and Systems},
  volume={6},
  pages={196--209},
  year={2024}
}

@inproceedings{an2024fluctuation,
  title={Fluctuation-based adaptive structured pruning for large language models},
  author={An, Yongqi and Zhao, Xu and Yu, Tao and Tang, Ming and Wang, Jinqiao},
  booktitle={Proceedings of the AAAI Conference on Artificial Intelligence},
  volume={38},
  number={10},
  pages={10865--10873},
  year={2024}
}

@article{zhu2024survey,
  title={A survey on model compression for large language models},
  author={Zhu, Xunyu and Li, Jian and Liu, Yong and Ma, Can and Wang, Weiping},
  journal={Transactions of the Association for Computational Linguistics},
  volume={12},
  pages={1556--1577},
  year={2024},
  publisher={MIT Press 255 Main Street, 9th Floor, Cambridge, Massachusetts 02142, USA~…}
}

@inproceedings{guo2017calibration,
  title={On calibration of modern neural networks},
  author={Guo, Chuan and Pleiss, Geoff and Sun, Yu and Weinberger, Kilian Q},
  booktitle={International conference on machine learning},
  pages={1321--1330},
  year={2017},
  organization={PMLR}
}

@inproceedings{gal2016dropout,
  title={Dropout as a bayesian approximation: Representing model uncertainty in deep learning},
  author={Gal, Yarin and Ghahramani, Zoubin},
  booktitle={international conference on machine learning},
  pages={1050--1059},
  year={2016},
  organization={PMLR}
}

@article{lakshminarayanan2017simple,
  title={Simple and scalable predictive uncertainty estimation using deep ensembles},
  author={Lakshminarayanan, Balaji and Pritzel, Alexander and Blundell, Charles},
  journal={Advances in neural information processing systems},
  volume={30},
  year={2017}
}

@article{kadavath2022language,
  title={Language models (mostly) know what they know},
  author={Kadavath, Saurav and Conerly, Tom and Askell, Amanda and Henighan, Tom and Drain, Dawn and Perez, Ethan and Schiefer, Nicholas and Hatfield-Dodds, Zac and DasSarma, Nova and Tran-Johnson, Eli and others},
  journal={arXiv preprint arXiv:2207.05221},
  year={2022}
}

@inproceedings{liu2025uncertainty,
  title={Uncertainty quantification and confidence calibration in large language models: A survey},
  author={Liu, Xiaoou and Chen, Tiejin and Da, Longchao and Chen, Chacha and Lin, Zhen and Wei, Hua},
  booktitle={Proceedings of the 31st ACM SIGKDD Conference on Knowledge Discovery and Data Mining V. 2},
  pages={6107--6117},
  year={2025}
}

@inproceedings{quach2024conformal,
  title={Conformal language modeling},
  author={Quach, Victor and Fisch, Adam and Schuster, Tal and Yala, Adam and Sohn, Jae Ho and Jaakkola, Tommi and Barzilay, Regina},
  booktitle={International Conference on Learning Representations},
  volume={2024},
  pages={11654--11681},
  year={2024}
}

@article{cherian2024large,
  title={Large language model validity via enhanced conformal prediction methods},
  author={Cherian, John J and Gibbs, Isaac and Cand{\`e}s, Emmanuel J},
  journal={Advances in Neural Information Processing Systems},
  volume={37},
  pages={114812--114842},
  year={2024}
}

@inproceedings{ulmer2024non,
  title={Non-exchangeable conformal language generation with nearest neighbors},
  author={Ulmer, Dennis and Zerva, Chrysoula and Martins, Andr{\'e} FT},
  booktitle={Findings of the Association for Computational Linguistics: EACL 2024},
  pages={1909--1929},
  year={2024}
}

@inproceedings{karimi2023quantifying,
  title={Quantifying deep learning model uncertainty in conformal prediction},
  author={Karimi, Hamed and Samavi, Reza},
  booktitle={Proceedings of the AAAI Symposium Series},
  volume={1},
  number={1},
  pages={142--148},
  year={2023}
}

@article{liang2022holistic,
  title={Holistic evaluation of language models},
  author={Liang, Percy and Bommasani, Rishi and Lee, Tony and Tsipras, Dimitris and Soylu, Dilara and Yasunaga, Michihiro and Zhang, Yian and Narayanan, Deepak and Wu, Yuhuai and Kumar, Ananya and others},
  journal={arXiv preprint arXiv:2211.09110},
  year={2022}
}

@article{chang2024survey,
  title={A survey on evaluation of large language models},
  author={Chang, Yupeng and Wang, Xu and Wang, Jindong and Wu, Yuan and Yang, Linyi and Zhu, Kaijie and Chen, Hao and Yi, Xiaoyuan and Wang, Cunxiang and Wang, Yidong and others},
  journal={ACM transactions on intelligent systems and technology},
  volume={15},
  number={3},
  pages={1--45},
  year={2024},
  publisher={ACM New York, NY}
}

@article{sensoy2018evidential,
  title={Evidential deep learning to quantify classification uncertainty},
  author={Sensoy, Murat and Kaplan, Lance and Kandemir, Melih},
  journal={Advances in neural information processing systems},
  volume={31},
  year={2018}
}

@article{kuhn2023semantic,
  title={Semantic uncertainty: Linguistic invariances for uncertainty estimation in natural language generation},
  author={Kuhn, Lorenz and Gal, Yarin and Farquhar, Sebastian},
  journal={arXiv preprint arXiv:2302.09664},
  year={2023}
}

@article{minderer2021revisiting,
  title={Revisiting the calibration of modern neural networks},
  author={Minderer, Matthias and Djolonga, Josip and Romijnders, Rob and Hubis, Frances and Zhai, Xiaohua and Houlsby, Neil and Tran, Dustin and Lucic, Mario},
  journal={Advances in neural information processing systems},
  volume={34},
  pages={15682--15694},
  year={2021}
}

@article{chang2026muoneq,
  title={Muoneq: Balancing before orthogonalization with lightweight equilibration},
  author={Chang, Da and Shi, Qiankun and Zhang, Lvgang and Li, Yu and Zhang, Ruijie and Lu, Yao and Liu, Yongxiang and Yuan, Ganzhao},
  journal={arXiv preprint arXiv:2603.28254},
  year={2026}
}

@article{chang2025convergence,
  title={On the Convergence of Muon and Beyond},
  author={Chang, Da and Liu, Yongxiang and Yuan, Ganzhao},
  journal={arXiv preprint arXiv:2509.15816},
  year={2025}
}

@article{chang2026mgup,
  title={MGUP: A Momentum-Gradient Alignment Update Policy for Stochastic Optimization},
  author={Chang, Da and Yuan, Ganzhao},
  journal={Advances in Neural Information Processing Systems},
  volume={38},
  pages={20488--20537},
  year={2026}
}

@article{li2026reason,
  title={Reason in Chains, Learn in Trees: Self-Rectification and Grafting for Multi-turn Agent Policy Optimization},
  author={Li, Yu and Tang, Sizhe and Lan, Tian},
  journal={arXiv preprint arXiv:2604.07165},
  year={2026}
}

@article{li2026craft,
  title={CRAFT-LoRA: Content-Style Personalization via Rank-Constrained Adaptation and Training-Free Fusion},
  author={Li, Yu and Cai, Yujun and Zhang, Chi},
  journal={arXiv preprint arXiv:2602.18936},
  year={2026}
}

@article{tong2025data,
  title={Data-Free Quantization of Vision Transformers via Easy-to-Hard Synthesis and Activation Correction},
  author={Tong, Yujia and Yuan, Jingling and Zhang, Tian and Liu, Jianquan and Hu, Chuang},
  journal={ACM Transactions on Multimedia Computing, Communications and Applications},
  year={2025},
  publisher={ACM New York, NY}
}

@article{zhang2026forget,
  title={Forget by Uncertainty: Orthogonal Entropy Unlearning for Quantized Neural Networks},
  author={Zhang, Tian and Tong, Yujia and Dong, Junhao and Xu, Ke and Wang, Yuze and Yuan, Jingling},
  journal={arXiv preprint arXiv:2602.00567},
  year={2026}
}

@article{tong2026sage,
  title={SAGE: Accelerating Vision-Language Models via Entropy-Guided Adaptive Speculative Decoding},
  author={Tong, Yujia and Zhang, Tian and Wan, Yunyang and Lin, Kaiwei and Yuan, Jingling and Hu, Chuang},
  journal={arXiv preprint arXiv:2602.00523},
  year={2026}
}

@article{guo2025optimal,
  title={Optimal brain restoration for joint quantization and sparsification of llms},
  author={Guo, Hang and Li, Yawei and Benini, Luca},
  journal={arXiv preprint arXiv:2509.11177},
  year={2025}
}

@InProceedings{Dong_2026_CVPR,
    author    = {Dong, Junhao and Zhang, Yifei and Zhu, Hao and Ong, Yew-Soon and Koniusz, Piotr},
    title     = {Hierarchically Robust Zero-shot Vision-language Models},
    booktitle = {Proceedings of the IEEE/CVF Conference on Computer Vision and Pattern Recognition (CVPR)},
    month     = {June},
    year      = {2026},
    pages     = {37642-37652}
}

@inproceedings{dong2026tug,
  title={Tug-of-war no more: Harmonizing accuracy and robustness in vision-language models via stability-aware task vector merging},
  author={Dong, Junhao and Qu, Xinghua and Zhang, Cong and Rong, Sua Qi and Thai, Nguyen Duc and Pan, Wenbo and Li, Xinfeng and Liu, Tongliang and Koniusz, Piotr and Ong, Yew-Soon},
  booktitle={The Fourteenth International Conference on Learning Representations},
  year={2026}
}

\clearpage
\appendix
\section{Appendix}

\subsection{Prompting Strategies}
\label{prompt}
Following~\cite{ye2024benchmarking}, we evaluate all models with prompt-based inference rather than task-specific finetuning. Since LLM outputs can be sensitive to prompt wording, we use three prompting strategies for every task: \textit{Base Prompt}, \textit{Shared Instruction Prompt}, and \textit{Task-specific Instruction Prompt}. The Base Prompt directly concatenates the question and all answer options, ending with an answer prefix. The Shared Instruction Prompt prepends a generic multiple-choice instruction, informing the model that exactly one option should be selected from six choices. The Task-specific Instruction Prompt instead uses an instruction tailored to the corresponding task, such as question answering, reading comprehension, commonsense inference, dialogue response selection, or document summarization.

For each prompted input, we obtain the logits of the option tokens $\{A,B,C,D,E,F\}$ at the final position and apply a softmax over these six logits to obtain the model's predictive distribution over answer choices. These probabilities are then used both for accuracy computation and for constructing conformal prediction sets. To reduce prompt-induced variance, all reported results are averaged over the three prompting strategies, together with the two conformal score functions described in Section~\ref{Per-Score}. Following the same setup, we include in-context demonstrations in the prompt: five demonstrations for QA, RC, and CI, three for DRS, and one for DS due to the longer input length of document summarization.

\subsection{Per-Score-Function Breakdown}
\label{Per-Score}

We further break down the progressive Wanda pruning results by conformal score function. Following~\cite{ye2024benchmarking}, we evaluate both Least Ambiguous Classifier (LAC) and Adaptive Prediction Set (APS). The two score functions use the same model predictions, so their Acc values are expected to be nearly identical, while their conformal set sizes can differ because LAC and APS construct prediction sets with different nonconformity scores.

Tables~\ref{tab:LAC} and~\ref{tab:APS} report the full per-task CR, Acc, and SS values for Llama2-7B and Llama3.1-8B under progressive Wanda pruning. Across both score functions, the same qualitative pattern holds: moderate pruning usually causes limited SS change, while higher sparsity, especially $40\%$--$50\%$, produces more visible uncertainty inflation on several tasks. APS often yields larger SS than LAC, especially for Llama3.1-8B, but both score functions support the threshold-like uncertainty behavior discussed in Section~\ref{sec:threshold}.

\begin{table*}[hptb]
\centering
\caption{Per-task results using the LAC score function under progressive Wanda pruning.}
\scriptsize
\label{tab:LAC}
\setlength{\tabcolsep}{2.4pt}
\begin{tabular}{llc ccc ccc ccc ccc ccc}
\toprule 
\multirow{2}{*}{\textbf{Model}} &
  \multirow{2}{*}{\textbf{Method}} &
  \multirow{2}{*}{\textbf{Sparsity}} &
  \multicolumn{3}{c}{\textbf{QA}} &
  \multicolumn{3}{c}{\textbf{RC}} &
  \multicolumn{3}{c}{\textbf{CI}} &
  \multicolumn{3}{c}{\textbf{DRS}} &
  \multicolumn{3}{c}{\textbf{DS}} \\
\cmidrule(lr){4-6} \cmidrule(lr){7-9} \cmidrule(lr){10-12} \cmidrule(lr){13-15} \cmidrule(lr){16-18}
& & & CR & Acc & SS & CR & Acc & SS & CR & Acc & SS & CR & Acc & SS & CR & Acc & SS \\
\midrule
\multirow{6}{*}{Llama2-7B}
  & Dense    & --   & 90.37 & 45.60 & 3.03 & 91.27 & 67.20 & 2.27 & 90.43 & 42.97 & 3.22 & 90.97 & 32.67 & 3.26 & 90.30 & 46.77 & 3.45\\
\cmidrule(l){2-18}
  & Wanda    & 10\% & 90.60 & 45.60 & 3.05 & 91.50 & 67.00 & 2.34 & 90.57 & 41.40 & 3.26 & 91.47 & 32.83 & 3.36 & 90.37 & 46.60 & 3.57\\
\cmidrule(l){2-18}
  & Wanda    & 20\% & 90.40 & 45.33 & 3.19 & 91.13 & 66.80 & 2.40 & 91.97 & 41.97 & 3.32 & 91.77 & 32.80 & 3.48 & 90.47 & 44.70 & 3.52\\
\cmidrule(l){2-18}
  & Wanda    & 30\% & 89.30 & 43.97 & 3.33 & 91.57 & 63.53 & 2.60 & 92.20 & 40.97 & 3.37 & 91.03 & 32.63 & 3.51 & 90.83 & 46.27 & 3.32\\
\cmidrule(l){2-18}
  & Wanda    & 40\% & 90.63 & 41.60 & 3.66 & 89.93 & 53.03 & 2.80 & 91.73 & 31.90 & 3.48 & 88.73 & 29.27 & 3.50 & 91.40 & 31.70 & 3.26\\
\cmidrule(l){2-18}
  & Wanda    & 50\% & 89.00 & 34.93 & 3.56 & 90.77 & 37.60 & 3.41 & 90.63 & 26.43 & 3.71 & 89.83 & 26.80 & 3.62 & 88.83 & 28.50 & 3.31\\
\midrule
\multirow{6}{*}{Llama3.1-8B}
  & Dense    & --   & 89.97 & 63.37 & 2.54 & 91.13 & 87.03 & 1.13 & 92.53 & 56.90 & 2.70 & 92.67 & 63.43 & 2.80 & 89.77 & 60.20 & 1.71\\
\cmidrule(l){2-18}
  & Wanda    & 10\% & 89.93 & 63.40 & 2.53 & 91.73 & 86.70 & 1.16 & 91.73 & 56.30 & 2.62 & 92.23 & 62.97 & 2.76 & 89.30 & 58.80 & 1.75\\
\cmidrule(l){2-18}
  & Wanda    & 20\% & 89.17 & 63.23 & 2.56 & 92.00 & 86.10 & 1.16 & 90.57 & 57.97 & 2.63 & 92.80 & 62.47 & 2.89 & 89.13 & 56.00 & 1.80\\
\cmidrule(l){2-18}
  & Wanda    & 30\% & 88.77 & 62.00 & 2.37 & 90.60 & 83.90 & 1.19 & 89.43 & 59.10 & 2.47 & 92.70 & 57.40 & 2.59 & 89.20 & 53.90 & 1.86\\
\cmidrule(l){2-18}
  & Wanda    & 40\% & 91.97 & 56.87 & 2.69 & 91.53 & 81.10 & 1.41 & 90.80 & 48.37 & 3.00 & 91.63 & 52.60 & 2.58 & 90.33 & 51.30 & 1.93\\
\cmidrule(l){2-18}
  & Wanda    & 50\% & 91.17 & 48.73 & 2.89 & 90.23 & 61.83 & 2.24 & 90.53 & 28.00 & 3.38 & 89.60 & 37.30 & 3.13 & 90.37 & 48.00 & 2.58\\
\bottomrule
\end{tabular}%
\end{table*}

\begin{table*}[hptb]
\centering
\caption{Per-task results using the APS score function under progressive Wanda pruning.}
\scriptsize
\label{tab:APS}
\setlength{\tabcolsep}{2.4pt}
\begin{tabular}{llc ccc ccc ccc ccc ccc}
\toprule 
\multirow{2}{*}{\textbf{Model}} &
  \multirow{2}{*}{\textbf{Method}} &
  \multirow{2}{*}{\textbf{Sparsity}} &
  \multicolumn{3}{c}{\textbf{QA}} &
  \multicolumn{3}{c}{\textbf{RC}} &
  \multicolumn{3}{c}{\textbf{CI}} &
  \multicolumn{3}{c}{\textbf{DRS}} &
  \multicolumn{3}{c}{\textbf{DS}} \\
\cmidrule(lr){4-6} \cmidrule(lr){7-9} \cmidrule(lr){10-12} \cmidrule(lr){13-15} \cmidrule(lr){16-18}
& & & CR & Acc & SS & CR & Acc & SS & CR & Acc & SS & CR & Acc & SS & CR & Acc & SS \\
\midrule
\multirow{6}{*}{Llama2-7B}
  & Dense    & --   & 92.43 & 45.46 & 3.37 & 92.39 & 67.20 & 2.65 & 90.51 & 43.57 & 3.30 & 90.49 & 32.67 & 3.34 & 89.40 & 46.77 & 3.03\\
\cmidrule(l){2-18}
  & Wanda    & 10\% & 92.20 & 45.86 & 3.41 & 92.34 & 67.20 & 2.68 & 90.37 & 41.34 & 3.38 & 90.63 & 32.91 & 3.40 & 88.69 & 46.26 & 3.13\\
\cmidrule(l){2-18}
  & Wanda    & 20\% & 91.36 & 45.27 & 3.47 & 91.57 & 66.60 & 2.64 & 90.77 & 42.17 & 3.40 & 91.13 & 32.60 & 3.46 & 89.17 & 44.64 & 3.26\\
\cmidrule(l){2-18}
  & Wanda    & 30\% & 91.90 & 43.77 & 3.59 & 91.93 & 63.73 & 2.86 & 89.90 & 41.03 & 3.37 & 89.63 & 32.31 & 3.51 & 90.61 & 46.13 & 3.22\\
\cmidrule(l){2-18}
  & Wanda    & 40\% & 90.87 & 41.54 & 3.78 & 89.93 & 52.63 & 2.98 & 89.13 & 31.90 & 3.40 & 89.87 & 29.33 & 3.60 & 90.54 & 31.90 & 3.16\\
\cmidrule(l){2-18}
  & Wanda    & 50\% & 89.80 & 36.21 & 3.62 & 91.73 & 37.60 & 3.55 & 91.43 & 26.71 & 3.83 & 88.77 & 27.34 & 3.60 & 91.47 & 29.16 & 3.59\\
\midrule
\multirow{6}{*}{Llama3.1-8B}
  & Dense    & --   & 93.47 & 62.49 & 3.42 & 98.57 & 87.11 & 2.65 & 91.77 & 58.30 & 2.90 & 92.47 & 63.17 & 2.56 & 90.63 & 60.74 & 2.07\\
\cmidrule(l){2-18}
  & Wanda    & 10\% & 94.03 & 63.66 & 3.49 & 97.67 & 86.30 & 2.66 & 91.87 & 55.30 & 2.90 & 92.91 & 64.09 & 2.74 & 90.44 & 60.46 & 2.03\\
\cmidrule(l){2-18}
  & Wanda    & 20\% & 93.59 & 62.97 & 3.72 & 97.34 & 86.96 & 2.72 & 91.03 & 57.77 & 2.87 & 92.46 & 62.73 & 2.61 & 91.33 & 63.00 & 2.10\\
\cmidrule(l){2-18}
  & Wanda    & 30\% & 95.69 & 60.80 & 3.53 & 97.00 & 83.50 & 2.83 & 89.33 & 60.10 & 2.79 & 92.90 & 59.66 & 2.73 & 92.60 & 57.90 & 2.22\\
\cmidrule(l){2-18}
  & Wanda    & 40\% & 94.17 & 57.93 & 3.31 & 96.61 & 80.90 & 2.71 & 91.26 & 55.77 & 3.08 & 91.87 & 49.80 & 2.50 & 89.03 & 45.76 & 2.25\\
\cmidrule(l){2-18}
  & Wanda    & 50\% & 90.29 & 43.21 & 3.29 & 90.57 & 56.97 & 2.82 & 90.77 & 35.34 & 3.46 & 89.56 & 33.70 & 3.19 & 89.43 & 43.94 & 2.46\\
\bottomrule
\end{tabular}%
\end{table*}

\subsection{Combined Quantization and Pruning}
\label{sec:combined}

\begin{table*}[t]
\centering
\caption{Results of combined quantization and pruning. The method applies OBR with W4A4K4 quantization and $50\%$ sparsity.}
\scriptsize
\label{tab:combined-quant-prune}
\setlength{\tabcolsep}{1.5pt}
\begin{tabular}{llc ccc ccc ccc ccc ccc}
\toprule
\multirow{2}{*}{\textbf{Model}} &
\multirow{2}{*}{\textbf{Method}} &
\multirow{2}{*}{\textbf{Sparsity}} &
\multicolumn{3}{c}{\textbf{QA}} &
\multicolumn{3}{c}{\textbf{RC}} &
\multicolumn{3}{c}{\textbf{CI}} &
\multicolumn{3}{c}{\textbf{DRS}} &
\multicolumn{3}{c}{\textbf{DS}} \\
\cmidrule(lr){4-6} \cmidrule(lr){7-9} \cmidrule(lr){10-12} \cmidrule(lr){13-15} \cmidrule(lr){16-18}
& & &
CR & Acc & SS &
CR & Acc & SS &
CR & Acc & SS &
CR & Acc & SS &
CR & Acc & SS \\
\midrule
Llama2-7B
& OBR(W4A4K4+$50\%$s)
& $50\%$
& 91.33 & 30.00 & 3.55
& 90.20 & 36.73 & 3.42
& 88.90 & 25.10 & 3.58
& 88.13 & 24.40 & 3.49
& 90.38 & 28.90 & 3.45 \\
Llama3.1-8B
& OBR(W4A4K4+$50\%$s)
& $50\%$
& 89.70 & 46.53 & 3.13
& 90.82 & 66.27 & 2.37
& 89.45 & 33.13 & 3.37
& 88.92 & 34.20 & 3.11
& 88.60 & 48.90 & 2.69 \\
\bottomrule
\end{tabular}
\end{table*}

We further evaluate a combined compression setting that applies quantization and pruning simultaneously, using OBR~\cite{guo2025optimal} with W4A4K4 quantization and $50\%$ sparsity. As shown in Table~\ref{tab:combined-quant-prune}, this combined setting substantially reduces accuracy on both evaluated models. For Llama2-7B, average Acc drops from the FP16 baseline of $47.09$ to $29.03$, while average SS increases from $3.09$ to $3.50$. For Llama3.1-8B, average Acc decreases from $66.27$ to $45.81$, and average SS increases from $2.45$ to $2.93$. Thus, combining low-bit quantization with high sparsity produces clear accuracy degradation, while the corresponding SS inflation remains moderate rather than catastrophic.

The task-level results again reveal accuracy--uncertainty decoupling. On Llama2-7B, RC accuracy drops sharply from $67.20\%$ to $36.73\%$, while SS increases from $2.46$ to $3.42$; on DRS, accuracy decreases from $32.67\%$ to $24.40\%$ with only a small SS increase from $3.30$ to $3.49$. A similar pattern appears for Llama3.1-8B: DRS accuracy drops from $63.30\%$ to $34.20\%$, but SS changes only from $2.68$ to $3.11$. Compared with standalone W4A4 quantization, the combined method does not always yield the largest SS inflation, but it consistently weakens predictive accuracy, reinforcing the need to evaluate both Acc and SS under joint compression.
\subsection{Per-Method Full Results Tables}

This section reports the complete per-task results for each compression method, model family, and model scale. For every task, we include Coverage Rate (CR), Prediction Accuracy (Acc), and Prediction Set Size (SS), allowing the aggregate trends discussed in the main text to be traced back to their task-level measurements. These tables also expose cases where accuracy and uncertainty move differently, such as settings with stable SS but large accuracy degradation, or near-baseline accuracy with noticeable SS inflation.

The tables are organized by compression paradigm and model family. Quantization results are grouped by bit-width and method, while pruning results are grouped by sparsity level and pruning algorithm. Unless otherwise noted, SS is the primary uncertainty metric, with larger values indicating larger prediction sets and therefore higher uncertainty. Entries marked as degenerate or with abnormal coverage should be interpreted cautiously, since their SS values may not reflect valid conformal uncertainty behavior.

\begin{table*}[hptb]
\centering
\caption{Comparison of Coverage Rate (CR \%), Prediction Accuracy (Acc \%), and Prediction Uncertainty (SS) across Llama2 models.}
\scriptsize
\label{tab:llama2-quant}
\setlength{\tabcolsep}{2.1pt}
\begin{tabular}{llc ccc ccc ccc ccc ccc}
\toprule \multirow{2}{*}{\textbf{Model}}
 &
  \multirow{2}{*}{\textbf{Bit}} &
  \multirow{2}{*}{\textbf{Method}} &
  \multicolumn{3}{c}{\textbf{QA}} &
  \multicolumn{3}{c}{\textbf{RC}} &
  \multicolumn{3}{c}{\textbf{CI}} &
  \multicolumn{3}{c}{\textbf{DRS}} &
  \multicolumn{3}{c}{\textbf{DS}} \\
  \cmidrule(lr){4-6} \cmidrule(lr){7-9} \cmidrule(lr){10-12} \cmidrule(lr){13-15} \cmidrule(lr){16-18}
  & & &
  CR & Acc & SS &
  CR & Acc & SS &
  CR & Acc & SS &
  CR & Acc & SS &
  CR & Acc & SS \\
\midrule
\multirow{7}{*}{Llama2-7B}
  & FP16   & --        &91.40 & 45.53 & 3.20 & 91.83 & 67.20 & 2.46 & 90.47 & 43.27 & 3.26 & 90.73 & 32.67 & 3.30 & 89.85 & 46.77 & 3.24  \\
  \noalign{\vspace{0.1em}}\cdashline{2-18}\noalign{\vspace{0.1em}}
  & W4A16  & RTN      &93.34&46.52&3.29&90.29&63.59&2.59&87.88&38.49&3.21&90.93&37.68&3.25&91.13&37.75&3.31  \\
  & W4A16  & AWQ       &91.30&45.92&3.18&91.37&63.45&2.66&87.38&43.84&3.18&91.57&30.92&3.47&88.05&31.79&3.45 \\
  & W4A16  & GPTQ      &91.48&43.05&3.22&89.51&63.83&2.35&91.65&43.05&3.31&91.00&33.90&3.30&91.95&47.16&3.35  \\
  \noalign{\vspace{0.1em}}\cdashline{2-18}\noalign{\vspace{0.1em}}
  & W4A4   & FlatQuant          &91.35&29.13&3.80&90.97&31.80&3.65&89.17&25.70&3.59&90.37&27.03&3.81&91.97&29.49&3.92 \\
  & W4A4   & QuaRot    &93.74&26.71&5.73&94.28&16.20&5.78&92.70&26.31&5.34&92.24&22.22&5.68&91.93&18.74&5.72 \\
  & W4A4   & SpinQuant &89.36&25.77&4.55&90.39&30.72&4.88&89.86&24.70&3.93&88.86&23.96&4.66&90.39&26.17&4.60 \\
\midrule
\multirow{7}{*}{Llama2-13B}
  & FP16   & --        &92.47 & 53.10 & 3.10 & 94.45 & 77.43 & 2.33 & 92.75 & 59.63 & 2.83 & 90.77 & 53.17 & 2.57 & 90.00 & 60.27 & 2.15\\
  \noalign{\vspace{0.1em}}\cdashline{2-18}\noalign{\vspace{0.1em}}
  & W4A16  & RTN       &92.27&52.73&3.11&93.58&75.30&2.58&91.55&56.77&3.13&92.28&49.12&2.99&90.33&63.33&2.13 \\
  & W4A16  & AWQ       &91.60&51.43&3.05&93.52&75.97&2.49&91.88&61.17&3.13&90.61&49.12&3.03&89.98&60.96&2.29\\
  & W4A16  & GPTQ      &91.47&52.23&3.00&93.87&76.60&2.30&92.38&58.33&2.88&91.71&49.22&2.82&90.10&57.58&2.43 \\
  \noalign{\vspace{0.1em}}\cdashline{2-18}\noalign{\vspace{0.1em}}
  & W4A4   &  FlatQuant         &91.53&44.07&3.25&91.77&62.60&2.74&89.85&38.30&3.33&89.42&39.34&3.14&90.18&48.13&2.96\\
  & W4A4   & QuaRot    &91.07&43.87&3.36&92.40&63.00&3.05&89.58&43.80&3.48&91.01&35.30&3.40&90.45&43.55&3.31 \\
  & W4A4   & SpinQuant &91.12&44.63&3.23&92.95&64.97&2.75&90.30&41.93&3.35&90.64&35.30&3.37&91.15&50.97&2.69\\
\midrule
\multirow{7}{*}{Llama2-70B}
  & FP16   & --        & 93.23 & 65.63 & 2.64 & 95.83 & 90.80 & 1.79 & 94.45 & 82.37 & 1.85 & 91.93 & 67.40 & 2.33 & 90.02 & 56.20 & 2.23\\
  \noalign{\vspace{0.1em}}\cdashline{2-18}\noalign{\vspace{0.1em}}
  & W4A16  & RTN       &92.45&64.40&2.56&95.52&89.57&1.92&94.28&83.03&1.86&92.23&66.23&2.45&91.40&54.24&2.34 \\
  & W4A16  & AWQ       &94.48&66.47&2.56&94.68&87.82&1.82&93.17&81.53&1.81&92.67&68.41&2.43&91.80&61.78&2.26 \\
  & W4A16  & GPTQ      &92.55 & 63.63 & 2.61 & 95.75 & 89.73 & 1.76 & 92.43 & 76.60 & 1.96 & 91.74 & 63.03 & 2.29 & 91.78 & 52.57 & 2.47\\
  \noalign{\vspace{0.1em}}\cdashline{2-18}\noalign{\vspace{0.1em}}
  & W4A4   &  FlatQuant         & 92.92&60.03&2.85&94.88&84.00&1.98&90.68&62.30&2.41&91.41&54.15&2.53&90.60&51.50&2.65
 \\
  & W4A4   & QuaRot    &93.40 & 59.83 & 2.80 & 94.80 & 84.37 & 2.11 & 93.10 & 76.97 & 2.04 & 91.22 & 59.56 & 2.80 & 90.63 & 51.74 & 2.95 \\
  & W4A4   & SpinQuant & 92.93&62.13&2.77&95.23&87.60&2.09&93.78&78.13&1.98&91.06&62.13&2.73&90.70&55.04&2.60 \\
\bottomrule
\end{tabular}%
\end{table*}

\begin{table*}[hptb]
\centering
\caption{Comparison of Coverage Rate (CR \%), Prediction Accuracy (Acc \%), and Prediction Uncertainty (SS) across other models.}
\scriptsize
\label{tab:compare-main-other}
\setlength{\tabcolsep}{2.4pt}
\begin{tabular}{llc ccc ccc ccc ccc ccc}
\toprule \multirow{2}{*}{\textbf{Method}}
 &
  \multirow{2}{*}{\textbf{Bit}} &
  \multirow{2}{*}{\textbf{Model}} &
  \multicolumn{3}{c}{\textbf{QA}} &
  \multicolumn{3}{c}{\textbf{RC}} &
  \multicolumn{3}{c}{\textbf{CI}} &
  \multicolumn{3}{c}{\textbf{DRS}} &
  \multicolumn{3}{c}{\textbf{DS}} \\
  \cmidrule(lr){4-6} \cmidrule(lr){7-9} \cmidrule(lr){10-12} \cmidrule(lr){13-15} \cmidrule(lr){16-18}
  & & &
  CR & Acc & SS &
  CR & Acc & SS &
  CR & Acc & SS &
  CR & Acc & SS &
  CR & Acc & SS \\
\midrule
\multirow{12}{*}{AWQ}
  & FP16   & DeepSeek-7B        &90.60 & 46.30 & 3.33 & 92.57 & 65.60 & 3.02 & 90.68 & 41.50 & 3.15 & 89.03 & 32.73 & 3.38 & 91.38 & 39.47 & 3.20 \\
  & W4A16  & DeepSeek-7B       &90.10 & 44.97 & 3.36 & 92.40 & 62.67 & 3.09 & 89.62 & 38.73 & 3.25 & 89.41 & 28.93 & 3.74 & 88.74 & 40.05 & 3.01 \\
\noalign{\vspace{0.1em}}\cdashline{2-18}
  & FP16   & Falcon-7B        &89.55 & 24.53 & 3.76 & 87.98 & 26.50 & 3.48 & 89.67 & 24.10 & 3.59 & 88.27 & 25.70 & 3.53 & 91.30 & 21.30 & 4.40 \\
  & W4A16  & Falcon-7B       &89.23 & 25.77 & 3.76 & 89.03 & 25.20 & 3.61 & 89.53 & 24.63 & 3.59 & 88.57 & 24.86 & 3.58 & 91.12 & 12.69 & 4.89 \\
\noalign{\vspace{0.1em}}\cdashline{2-18}\noalign{\vspace{0.1em}}
 & FP16   & Llama2-7B        &91.40 & 45.53 & 3.20 & 91.83 & 67.20 & 2.46 & 90.47 & 43.27 & 3.26 & 90.73 & 32.67 & 3.30 & 89.85 & 46.77 & 3.24\\
  & W4A16  & Llama2-7B       &91.30&45.92&3.18&91.37&63.45&2.66&87.38&43.84&3.18&91.57&30.92&3.47&88.05&31.79&3.45\\
\noalign{\vspace{0.1em}}\cdashline{2-18}\noalign{\vspace{0.1em}}
 & FP16   & Llama3.1-8B        &91.72 & 62.93 & 2.98 & 94.85 & 87.07 & 1.89 & 92.15 & 57.60 & 2.80 & 92.57 & 63.30 & 2.68 & 90.20 & 60.47 & 1.89\\
  & W4A16  & Llama3.1-8B       &91.83 & 58.25 & 2.15 & 92.03 & 64.17 & 2.95 & 91.86 & 65.77 & 2.64 & 94.30 & 87.63 & 1.95 & 91.13 & 62.20 & 2.95\\
\noalign{\vspace{0.1em}}\cdashline{2-18}\noalign{\vspace{0.1em}}
  & FP16   & Qwen3-8B    &94.20 & 70.43 & 3.00 & 93.70 & 84.17 & 2.20 & 94.67 & 82.10 & 2.31 & 90.40 & 58.03 & 3.35 & 90.57 & 57.47 & 2.56\\
  & W4A16   & Qwen3-8B &90.48 & 55.41 & 2.64 & 94.43 & 72.03 & 3.11 & 92.33 & 57.12 & 3.07 & 93.85 & 83.00 & 2.27 & 94.03 & 80.47 & 2.31 \\
\midrule
\multirow{8}{*}{QuaRot}
  & FP16   & DeepSeek-7B        &90.60 & 46.30 & 3.33 & 92.57 & 65.60 & 3.02 & 90.68 & 41.50 & 3.15 & 89.03 & 32.73 & 3.38 & 91.38 & 39.47 & 3.20 \\
  & W4A4  & DeepSeek-7B       &90.13&34.90&3.79&90.82&45.83&3.41&89.82&31.40&3.61&88.96&27.99&4.14&89.11&35.17&3.67 \\
\noalign{\vspace{0.1em}}\cdashline{2-18}\noalign{\vspace{0.1em}}
 & FP16   & Llama2-7B        &91.40 & 45.53 & 3.20 & 91.83 & 67.20 & 2.46 & 90.47 & 43.27 & 3.26 & 90.73 & 32.67 & 3.30 & 89.85 & 46.77 & 3.24 \\
  & W4A4  & Llama2-7B       &93.74&26.71&5.73&94.28&16.20&5.78&92.70&26.31&5.34&92.24&22.22&5.68&91.93&18.74&5.72\\
\noalign{\vspace{0.1em}}\cdashline{2-18}\noalign{\vspace{0.1em}}
 & FP16   & Llama3.1-8B        &91.72 & 62.93 & 2.98 & 94.85 & 87.07 & 1.89 & 92.15 & 57.60 & 2.80 & 92.57 & 63.30 & 2.68 & 90.20 & 60.47 & 1.89\\
  & W4A4  & Llama3.1-8B       &89.75 & 37.74 & 3.27 & 90.87 & 47.63 & 3.28 & 89.41 & 38.37 & 3.47 & 91.50 & 65.53 & 2.48 & 90.27 & 39.70 & 3.49\\
\noalign{\vspace{0.1em}}\cdashline{2-18}\noalign{\vspace{0.1em}}
  & FP16   & Qwen3-8B    &94.20 & 70.43 & 3.00 & 93.70 & 84.17 & 2.20 & 94.67 & 82.10 & 2.31 & 90.40 & 58.03 & 3.35 & 90.57 & 57.47 & 2.56\\
  & W4A4   & Qwen3-8B &92.67&14.46&5.51&92.14&18.67&5.44&91.77&18.88&5.50&93.04&19.01&5.50&91.37&20.28&5.36 \\
\midrule
\end{tabular}%
\end{table*}

\begin{table*}[hptb]
\centering
\caption{Comparison of Coverage Rate (CR \%), Prediction Accuracy (Acc \%), and Prediction Uncertainty (SS) across Llama3 models.}
\scriptsize
\label{tab:compare-main-llama3-combined}
\setlength{\tabcolsep}{2.4pt}
\begin{tabular}{llc ccc ccc ccc ccc ccc}
\toprule \multirow{2}{*}{\textbf{Model}}
 &
  \multirow{2}{*}{\textbf{Bit}} &
  \multirow{2}{*}{\textbf{Method}} &
  \multicolumn{3}{c}{\textbf{QA}} &
  \multicolumn{3}{c}{\textbf{RC}} &
  \multicolumn{3}{c}{\textbf{CI}} &
  \multicolumn{3}{c}{\textbf{DRS}} &
  \multicolumn{3}{c}{\textbf{DS}} \\
  \cmidrule(lr){4-6} \cmidrule(lr){7-9} \cmidrule(lr){10-12} \cmidrule(lr){13-15} \cmidrule(lr){16-18}
  & & &
  CR & Acc & SS &
  CR & Acc & SS &
  CR & Acc & SS &
  CR & Acc & SS &
  CR & Acc & SS \\
\midrule
\multirow{7}{*}{Llama3.2-1B}
  & FP16   & --        & 92.02 & 33.87 & 3.55 & 90.12 & 38.27 & 3.43 & 88.82 & 24.13 & 3.69 & 88.50 & 24.67 & 3.54 & 92.03 & 24.37 & 3.62 \\
  \noalign{\vspace{0.1em}}\cdashline{2-18}\noalign{\vspace{0.1em}}
  & W4A16  & RTN       &91.83 & 32.30 & 3.62 & 89.62 & 32.27 & 3.53 & 88.18 & 25.00 & 3.61 & 88.36 & 23.56 & 3.58 & 90.16 & 22.34 & 3.65 \\
  & W4A16  & AWQ       &91.98 & 30.83 & 3.57 & 89.95 & 28.70 & 3.43 & 89.85 & 25.97 & 3.60 & 87.79 & 23.72 & 3.57 & 89.13 & 20.37 & 4.65 \\
  & W4A16  & GPTQ      &91.17&29.27&3.57&90.07&26.53&3.50&89.75&24.37&3.66&88.17&23.46&3.58&89.20&25.45&3.75 \\
  \noalign{\vspace{0.1em}}\cdashline{2-18}\noalign{\vspace{0.1em}}
  & W4A4   & FlatQuant          &89.18&24.77&4.18&88.18&23.60&4.50&89.77&23.17&4.13&88.56&24.39&4.46&88.61&24.52&4.75
 \\
  & W4A4   & QuaRot    &92.57&23.09&5.70&93.21&14.73&5.40&92.50&16.87&5.38&92.07&28.05&5.67&93.24&19.34&5.41\\
  & W4A4   & SpinQuant &88.73 & 25.70 & 3.74 & 89.78 & 25.73 & 3.80 & 89.83 & 25.23 & 3.66 & 90.16 & 24.06 & 3.94 & 91.15 & 23.18 & 4.78\\
\midrule
\multirow{7}{*}{Llama3.1-8B}
  & FP16   & --        &91.72 & 62.93 & 2.98 & 94.85 & 87.07 & 1.89 & 92.15 & 57.60 & 2.80 & 92.57 & 63.30 & 2.68 & 90.20 & 60.47 & 1.89 \\
  \noalign{\vspace{0.1em}}\cdashline{2-18}\noalign{\vspace{0.1em}}
  & W4A16  & RTN       &91.72 & 63.00 & 2.98 & 94.85 & 87.03 & 1.90 & 92.25 & 57.57 & 2.81 & 92.18 & 62.83 & 2.64 & 91.62 & 61.46 & 1.95 \\
  & W4A16  & AWQ       &92.03 & 64.17 & 2.95 & 94.30 & 87.63 & 1.95 & 91.13 & 62.20 & 2.95 & 91.86 & 65.77 & 2.64 & 91.83 & 58.25 & 2.15\\
  & W4A16  & GPTQ      &93.98&63.12&2.78&94.11&83.94&2.00&87.78&53.82&2.86&91.87&64.32&2.27&90.76&66.53&1.84\\
  \noalign{\vspace{0.1em}}\cdashline{2-18}\noalign{\vspace{0.1em}}
  & W4A4   &  FlatQuant         &91.40&35.61&4.06&92.37&42.77&3.87&91.16&30.12&4.10&90.23&35.74&3.34&88.79&36.01&3.15\\
  & W4A4   & QuaRot    &90.87 & 47.63 & 3.28 & 91.50 & 65.53 & 2.48 & 90.27 & 39.70 & 3.49 & 89.41 & 38.37 & 3.47 & 89.75 & 37.74 & 3.27\\
  & W4A4   & SpinQuant &92.27&51.07&3.17&92.37&68.27&2.41&87.08&41.57&3.28&91.50&40.63&3.20&92.27&46.59&3.03 \\
\midrule
\multirow{7}{*}{Llama3.1-70B}
  & FP16   & --        & 92.53 & 76.20 & 2.17 & 96.32 & 93.13 & 1.61 & 94.43 & 81.33 & 1.89 & 91.23 & 74.97 & 1.88 & 90.95 & 62.37 & 2.06\\
  \noalign{\vspace{0.1em}}\cdashline{2-18}\noalign{\vspace{0.1em}}
  & W4A16  & RTN       &92.40&74.70&2.32&94.67&89.03&1.75&92.78&66.73&2.37&93.89&73.27&2.08&91.23&54.71&2.18 \\
  & W4A16  & AWQ       & 92.90 & 74.00 & 2.27 & 95.58 & 91.67 & 1.66 & 93.25 & 77.40 & 2.06 & 91.47 & 73.24 & 2.10 & 91.77 & 59.25 & 2.61
  \\
  & W4A16  & GPTQ      &92.73 & 75.07 & 2.28 & 96.17 & 92.67 & 1.70 & 94.52 & 80.60 & 1.88 & 92.71 & 75.48 & 1.96 & 91.48 & 65.30 & 1.98 \\
  \noalign{\vspace{0.1em}}\cdashline{2-18}\noalign{\vspace{0.1em}}
  & W4A4   &  FlatQuant   & -- & -- & -- & -- & -- &  -- & -- & -- & -- & -- & -- & -- & -- & -- & -- \\
  & W4A4   & QuaRot    & -- & -- & -- & -- & -- &  -- & -- & -- & -- & -- & -- & -- & -- & -- & --\\
  & W4A4   & SpinQuant & -- & -- & -- & -- & -- &  -- & -- & -- & -- & -- & -- & -- & -- & -- & --\\
\bottomrule
\end{tabular}%
\end{table*}

\begin{table*}[hptb]
\centering
\caption{Comparison of Coverage Rate (CR \%), Prediction Accuracy (Acc \%), and Prediction Uncertainty (SS) across Llama2 models.}
\scriptsize
\label{tab:llama2-pruning}
\setlength{\tabcolsep}{2.2pt}
\begin{tabular}{llc ccc ccc ccc ccc ccc}
\toprule 
\multirow{2}{*}{\textbf{Model}} &
  \multirow{2}{*}{\textbf{Method}} &
  \multirow{2}{*}{\textbf{Sparsity}} &
  \multicolumn{3}{c}{\textbf{QA}} &
  \multicolumn{3}{c}{\textbf{RC}} &
  \multicolumn{3}{c}{\textbf{CI}} &
  \multicolumn{3}{c}{\textbf{DRS}} &
  \multicolumn{3}{c}{\textbf{DS}} \\
\cmidrule(lr){4-6} \cmidrule(lr){7-9} \cmidrule(lr){10-12} \cmidrule(lr){13-15} \cmidrule(lr){16-18}
& & & CR & Acc & SS & CR & Acc & SS & CR & Acc & SS & CR & Acc & SS & CR & Acc & SS \\
\midrule
%% ---------- Llama2-7B ----------
\multirow{9}{*}{Llama2-7B} 
  & Dense    & --   & 91.40 & 45.53 & 3.20 & 91.83 & 67.20 & 2.46 & 90.47 & 43.27 & 3.26 & 90.73 & 32.67 & 3.30 & 89.85 & 46.77 & 3.24\\
\cmidrule(l){2-18}
  & \multicolumn{17}{c}{\textit{Unstructured Pruning}} \\
\cmidrule(l){2-18}
  & Magnitude & 50\% & 90.25 & 29.67 & 4.13 & 90.70 & 28.13 & 4.91 & 92.00 & 25.03 & 4.37 & 90.03 & 23.23 & 4.61 & 88.17 & 18.83 & 4.73 \\
  & SparseGPT& 50\% & 89.25 & 35.90 & 3.67 & 89.85 & 53.60 & 2.99 & 90.17 & 27.83 & 3.50 & 89.17 & 24.70 & 3.58 & 90.47 & 28.60 & 3.65\\
  & Wanda  & 50\% & 89.40 & 35.57 & 3.59 & 91.25 & 37.60 & 3.48 & 91.03 & 26.57 & 3.77 & 89.30 & 27.07 & 3.61 & 90.15 & 28.83 & 3.45 \\
\cmidrule(l){2-18}
  & \multicolumn{17}{c}{\textit{Structured Pruning}} \\
\cmidrule(l){2-18}
  & LLM-Pruner   & 20\% &  93.55 & 25.00 & 5.74 & 93.33 & 25.50 & 5.74 & 93.05 & 23.50 & 5.73 & 93.72 & 25.70 & 5.74 & 93.80 & 26.40 & 5.76 \\
  & SliceGPT     & 20\% &  91.95 & 28.17 & 3.66 & 90.75 & 31.03 & 3.32 & 89.77 & 24.03 & 3.63 & 90.60 & 24.40 & 3.63 & 90.88 & 24.53 & 4.46 \\
\midrule
%% ---------- Llama2-13B ----------
\multirow{9}{*}{Llama2-13B}
  & Dense    & --   & 92.47 & 53.10 & 3.10 & 94.45 & 77.43 & 2.33 & 92.75 & 59.63 & 2.83 & 90.77 & 53.17 & 2.57 & 90.00 & 60.27 & 2.15 \\
\cmidrule(l){2-18}
  & \multicolumn{17}{c}{\textit{Unstructured Pruning}} \\
\cmidrule(l){2-18}
  & Magnitude & 50\% & 90.55 & 39.93 & 3.21 & 92.92 & 59.20 & 2.75 & 89.77 & 35.27 & 3.49 & 88.50 & 39.00 & 3.07 & 91.97 & 46.37 & 3.00\\
  & SparseGPT& 50\% & 90.92 & 47.23 & 3.15 & 92.20 & 69.60 & 2.26 & 90.40 & 34.37 & 3.34 & 88.93 & 38.60 & 3.33 & 90.02 & 40.80 & 3.10 \\
  & Wanda     & 50\% & 91.77 & 45.57 & 3.20 & 92.50 & 66.20 & 2.56 & 89.20 & 39.83 & 3.46 & 89.73 & 40.83 & 3.21 & 90.55 & 53.10 & 2.44 \\
\cmidrule(l){2-18}
  & \multicolumn{17}{c}{\textit{Structured Pruning}} \\
\cmidrule(l){2-18}
  & LLM-Pruner   & 20\% & 91.55 & 43.43 & 3.45 & 92.57 & 53.33 & 3.29 & 92.38 & 26.67 & 4.42 & 90.00 & 16.30 & 4.65 & 88.12 & 20.17 & 4.54 \\
  & SliceGPT     & 20\% & 90.30 & 40.93 & 3.30 & 91.83 & 63.30 & 2.78 & 90.10 & 26.83 & 4.38 & 88.97 & 30.67 & 4.08 & 91.80 & 33.37 & 3.71 \\
\midrule
%% ---------- Llama2-70B ----------
\multirow{9}{*}{Llama2-70B}
  & Dense    & --   & 93.23 & 65.63 & 2.64 & 95.83 & 90.80 & 1.79 & 94.45 & 82.37 & 1.85 & 91.93 & 67.40 & 2.33 & 90.02 & 56.20 & 2.23\\
\cmidrule(l){2-18}
  & \multicolumn{17}{c}{\textit{Unstructured Pruning}} \\
\cmidrule(l){2-18}
  & Magnitude & 50\% & 92.20 & 56.90 & 3.25 & 93.92 & 79.90 & 2.29 & 92.63 & 56.33 & 2.44 & 90.33 & 55.03 & 2.88 & 90.25 & 55.03 & 2.80\\
  & SparseGPT & 50\% &  92.62 & 60.50 & 2.82 & 95.32 & 86.57 & 1.98 & 91.35 & 69.40 & 2.17 & 91.03 & 58.37 & 2.68 & 90.38 & 51.67 & 2.53 \\
  & Wanda    & 50\% & 91.95 & 60.53 & 2.85 & 95.35 & 87.97 & 1.94 & 92.75 & 72.57 & 2.14 & 90.45 & 59.53 & 2.60 & 91.13 & 52.53 & 2.55 \\
\cmidrule(l){2-18}
  & \multicolumn{17}{c}{\textit{Structured Pruning}} \\
\cmidrule(l){2-18}
  & LLM-Pruner   & 20\% & 93.18 & 62.77 & 2.96 & 95.53 & 88.07 & 2.03 & 94.67 & 78.77 & 2.06 & 91.67 & 57.37 & 3.32 & 90.30 & 49.97 & 2.62 \\
  & SliceGPT     & 20\% & 92.03 & 59.30 & 2.84 & 93.92 & 84.13 & 1.85 & 90.15 & 46.67 & 2.90 & 90.22 & 51.50 & 3.11 & 91.63 & 45.03 & 2.81 \\
\bottomrule
\end{tabular}%
\end{table*}

\begin{table*}[hptb]
\centering
\caption{Comparison of Coverage Rate (CR \%), Prediction Accuracy (Acc \%), and Prediction Uncertainty (SS) across Qwen3(Dense) models.}
\scriptsize
\label{tab:compare-main-qwen3-dense-combined}
\setlength{\tabcolsep}{2.2pt}
\begin{tabular}{llc ccc ccc ccc ccc ccc}
\toprule 
\multirow{2}{*}{\textbf{Model}} &
  \multirow{2}{*}{\textbf{Method}} &
  \multirow{2}{*}{\textbf{Sparsity}} &
  \multicolumn{3}{c}{\textbf{QA}} &
  \multicolumn{3}{c}{\textbf{RC}} &
  \multicolumn{3}{c}{\textbf{CI}} &
  \multicolumn{3}{c}{\textbf{DRS}} &
  \multicolumn{3}{c}{\textbf{DS}} \\
\cmidrule(lr){4-6} \cmidrule(lr){7-9} \cmidrule(lr){10-12} \cmidrule(lr){13-15} \cmidrule(lr){16-18}
& & & CR & Acc & SS & CR & Acc & SS & CR & Acc & SS & CR & Acc & SS & CR & Acc & SS \\
\midrule
%% ---------- Qwen3-8B ----------
\multirow{9}{*}{Qwen3-8B}
  & Dense     & --   &94.20 & 70.43 & 3.00 & 93.70 & 84.17 & 2.20 & 94.67 & 82.10 & 2.31 & 90.40 & 58.03 & 3.35 & 90.57 & 57.47 & 2.56\\
\cmidrule(l){2-18}
  & \multicolumn{17}{c}{\textit{Unstructured Pruning}} \\
\cmidrule(l){2-18}
  & Magnitude& 50\% & 91.03 & 38.27 & 3.33 & 91.73 & 32.90 & 3.60 & 90.38 & 26.93 & 3.49 & 89.40 & 27.23 & 3.58 & 90.37 & 23.63 & 3.71\\
  & SparseGPT& 50\% &93.22 & 63.77 & 3.14 & 93.78 & 81.80 & 2.22 & 93.45 & 77.33 & 2.02 & 91.32 & 52.60 & 3.31 & 90.68 & 44.97 & 2.96\\
  & Wanda    & 50\% &  93.52 & 65.37 & 3.30 & 93.33 & 81.87 & 2.21 & 92.77 & 68.73 & 2.35 & 92.88 & 53.80 & 3.34 & 91.88 & 46.97 & 2.93\\
\cmidrule(l){2-18}
  & \multicolumn{17}{c}{\textit{Structured Pruning}} \\
\cmidrule(l){2-18}
  & LLM-Pruner   & 20\% &  94.15 & 64.00 & 3.24 & 93.73 & 82.97 & 2.17 & 92.50 & 67.50 & 2.54 & 91.15 & 53.67 & 3.27 & 91.38 & 54.57 & 2.43 \\
  & SliceGPT     & 20\% & 12.50 & 25.00 & 1.00 & 12.75 & 25.50 & 1.00 & 11.75 & 23.50 & 1.00 & 12.85 & 25.70 & 1.00 & 13.20 & 26.40 & 1.00 \\
\midrule
%% ---------- Qwen3-14B ----------
\multirow{9}{*}{Qwen3-14B}
  & Dense    & --   & 94.02 & 74.33 & 2.71 & 94.38 & 84.97 & 2.09 & 94.83 & 86.60 & 1.86 & 91.12 & 56.10 & 3.16 & 89.48 & 60.50 & 2.25\\
\cmidrule(l){2-18}
  & \multicolumn{17}{c}{\textit{Unstructured Pruning}} \\
\cmidrule(l){2-18}
  & Magnitude & 50\% & 92.77 & 62.53 & 2.81 & 92.47 & 73.33 & 2.43 & 92.40 & 48.83 & 2.61 & 90.22 & 55.83 & 2.70 & 91.32 & 56.07 & 2.32 \\
  & SparseGPT & 50\% & 93.18 & 71.80 & 2.48 & 93.77 & 84.30 & 2.01 & 94.98 & 83.57 & 1.91 & 92.35 & 62.13 & 2.83 & 92.20 & 55.10 & 2.24\\
  & Wanda    & 50\% &  94.02 & 71.77 & 2.51 & 94.53 & 85.50 & 2.02 & 94.33 & 84.13 & 1.81 & 90.03 & 60.30 & 2.63 & 90.85 & 58.33 & 2.02\\
\cmidrule(l){2-18}
  & \multicolumn{17}{c}{\textit{Structured Pruning}} \\
\cmidrule(l){2-18}
  & LLM-Pruner   & 20\% & 91.37 & 67.63 & 3.02 & 93.40 & 79.47 & 2.12 & 93.43 & 80.13 & 2.29 & 91.70 & 61.00 & 2.97 & 89.98 & 55.80 & 2.37 \\
  & SliceGPT     & 20\% & 91.95 & 20.13 & 5.30 & 91.12 & 18.07 & 5.27 & 90.17 & 19.60 & 5.10 & 90.65 & 18.03 & 5.22 & 91.60 & 17.97 & 5.29 \\
\midrule
%% ---------- Qwen3-32B ----------
\multirow{9}{*}{Qwen3-32B}
  & Dense    & --   &  65.18 & 58.63 & 2.06 & 93.23 & 83.97 & 1.93 & 93.83 & 86.97 & 1.73 & 90.95 & 62.03 & 3.31 & 80.90 & 64.03 & 1.79\\
\cmidrule(l){2-18}
  & \multicolumn{17}{c}{\textit{Unstructured Pruning}} \\
\cmidrule(l){2-18}
  & Magnitude& 50\% & 65.92 & 50.90 & 2.49 & 92.07 & 72.33 & 2.34 & 90.53 & 70.10 & 2.33 & 90.50 & 53.60 & 3.37 & 83.68 & 67.70 & 1.61\\
  & SparseGPT & 50\% & 66.28 & 57.27 & 2.33 & 93.68 & 87.33 & 1.86 & 93.43 & 83.73 & 2.03 & 91.33 & 62.23 & 3.32 & 79.92 & 59.77 & 2.03 \\
  & Wanda    & 50\% &  64.63 & 54.90 & 2.40 & 93.52 & 86.53 & 1.88 & 93.97 & 83.40 & 2.18 & 91.50 & 61.73 & 3.55 & 80.65 & 62.33 & 1.73\\
\cmidrule(l){2-18}
  & \multicolumn{17}{c}{\textit{Structured Pruning}} \\
\cmidrule(l){2-18}
  & LLM-Pruner   & 20\% & 65.70 & 55.40 & 2.38 & 93.40 & 81.37 & 1.96 & 93.65 & 82.83 & 1.98 & 89.75 & 59.43 & 3.05 & 76.98 & 54.10 & 2.09 \\
  & SliceGPT     & 20\% &  65.60 & 23.13 & 3.77 & 89.73 & 21.10 & 5.07 & 88.80 & 21.53 & 5.03 & 90.83 & 19.90 & 5.13 & 66.85 & 18.87 & 3.86 \\
\bottomrule
\end{tabular}%
\end{table*}

\begin{table*}[hptb]
\centering
\caption{Comparison of Coverage Rate (CR \%), Prediction Accuracy (Acc \%), and Prediction Uncertainty (SS) across Llama3 models.}
\scriptsize
\label{tab:llama3-pruning}
\setlength{\tabcolsep}{2.1pt}
\begin{tabular}{llc ccc ccc ccc ccc ccc}
\toprule 
\multirow{2}{*}{\textbf{Model}} &
  \multirow{2}{*}{\textbf{Method}} &
  \multirow{2}{*}{\textbf{Sparsity}} &
  \multicolumn{3}{c}{\textbf{QA}} &
  \multicolumn{3}{c}{\textbf{RC}} &
  \multicolumn{3}{c}{\textbf{CI}} &
  \multicolumn{3}{c}{\textbf{DRS}} &
  \multicolumn{3}{c}{\textbf{DS}} \\
\cmidrule(lr){4-6} \cmidrule(lr){7-9} \cmidrule(lr){10-12} \cmidrule(lr){13-15} \cmidrule(lr){16-18}
& & & CR & Acc & SS & CR & Acc & SS & CR & Acc & SS & CR & Acc & SS & CR & Acc & SS \\
\midrule
%% ---------- Llama3.2-1B ----------
\multirow{9}{*}{Llama3.2-1B}
  & Dense    & --   & 92.02 & 33.87 & 3.55 & 90.12 & 38.27 & 3.43 & 88.82 & 24.13 & 3.69 & 88.50 & 24.67 & 3.54 & 92.03 & 24.37 & 3.62\\
\cmidrule(l){2-18}
  & \multicolumn{17}{c}{\textit{Unstructured Pruning}} \\
\cmidrule(l){2-18}
  & Magnitude & 50\% &90.17 & 25.07 & 3.61 & 88.55 & 24.70 & 3.58 & 90.53 & 23.50 & 5.10 & 91.60 & 24.73 & 4.15 & 90.67 & 24.57 & 4.79\\
  & SparseGPT & 50\% & 88.75 & 24.93 & 3.54 & 88.13 & 25.03 & 3.50 & 90.92 & 24.10 & 3.62 & 90.77 & 25.27 & 3.63 & 89.45 & 26.97 & 3.63\\
  & Wanda    & 50\% & 89.48 & 24.67 & 3.60 & 89.08 & 25.47 & 3.56 & 89.65 & 24.70 & 3.61 & 89.27 & 25.70 & 3.67 & 91.15 & 25.87 & 3.81\\
\cmidrule(l){2-18}
  & \multicolumn{17}{c}{\textit{Structured Pruning}} \\
\cmidrule(l){2-18}
  & LLM-Pruner   & 20\% &  90.83 & 26.60 & 5.00 & 92.67 & 25.97 & 5.54 & 91.78 & 23.93 & 3.76 & 89.33 & 23.13 & 4.15 & 88.32 & 21.00 & 4.51 \\
  & SliceGPT     & 20\% & 90.58 & 25.13 & 3.64 & 88.72 & 25.10 & 3.56 & 88.75 & 24.10 & 3.60 & 88.35 & 25.70 & 4.94 & 90.32 & 24.33 & 4.07 \\
\midrule
%% ---------- Llama3.1-8B ----------
\multirow{9}{*}{Llama3.1-8B}
  & Dense    & --   &  91.72 & 62.93 & 2.98 & 94.85 & 87.07 & 1.89 & 92.15 & 57.60 & 2.80 & 92.57 & 63.30 & 2.68 & 90.20 & 60.47 & 1.89 \\
\cmidrule(l){2-18}
  & \multicolumn{17}{c}{\textit{Unstructured Pruning}} \\
\cmidrule(l){2-18}
  & Magnitude & 50\% & 91.27 & 31.30 & 4.02 & 89.65 & 27.00 & 3.71 & 90.75 & 24.33 & 3.76 & 91.33 & 23.50 & 3.72 & 93.65 & 23.00 & 5.71\\
  & SparseGPT & 50\% &   91.20 & 47.20 & 3.06 & 91.95 & 73.20 & 2.17 & 91.08 & 37.63 & 3.41 & 90.35 & 35.57 & 3.26 & 93.03 & 42.77 & 3.13\\
  & Wanda    & 50\% &90.73 & 45.97 & 3.09 & 90.40 & 59.40 & 2.53 & 90.65 & 31.67 & 3.42 & 89.58 & 35.50 & 3.16 & 89.90 & 45.97 & 2.52\\
\cmidrule(l){2-18}
  & \multicolumn{17}{c}{\textit{Structured Pruning}} \\
\cmidrule(l){2-18}
  & LLM-Pruner   & 20\% & 93.17 & 52.03 & 3.03 & 90.43 & 67.63 & 2.29 & 89.82 & 46.97 & 3.28 & 89.40 & 37.33 & 3.31 & 89.53 & 37.83 & 3.29 \\
  & SliceGPT     & 20\% &  90.97 & 40.87 & 3.44 & 89.95 & 60.13 & 3.12 & 89.55 & 27.87 & 3.60 & 90.18 & 25.73 & 3.60 & 91.68 & 28.20 & 3.89 \\
\midrule
%% ---------- Llama3.1-70B ----------
\multirow{9}{*}{Llama3.1-70B}
  & Dense    & --   & 92.53 & 76.20 & 2.17 & 96.32 & 93.13 & 1.61 & 94.43 & 81.33 & 1.89 & 91.23 & 74.97 & 1.88 & 90.95 & 62.37 & 2.06 \\
\cmidrule(l){2-18}
  & \multicolumn{17}{c}{\textit{Unstructured Pruning}} \\
\cmidrule(l){2-18}
  & Magnitude& 50\% & 92.95 & 63.17 & 2.90 & 92.62 & 77.73 & 2.09 & 89.85 & 46.47 & 3.25 & 87.85 & 43.37 & 3.78 & 89.98 & 48.47 & 2.87 \\
  & SparseGPT& 50\% &  94.30 & 71.10 & 2.40 & 94.67 & 89.33 & 1.65 & 90.30 & 60.83 & 2.79 & 90.07 & 57.67 & 2.44 & 91.67 & 47.07 & 2.36 \\
  & Wanda  & 50\% & 93.65 & 71.37 & 2.33 & 94.33 & 88.83 & 1.56 & 90.80 & 49.53 & 3.16 & 89.05 & 58.80 & 2.32 & 90.88 & 55.97 & 2.07\\
\cmidrule(l){2-18}
  & \multicolumn{17}{c}{\textit{Structured Pruning}} \\
\cmidrule(l){2-18}
  & LLM-Pruner   & 20\% & 94.02 & 69.43 & 2.60 & 94.82 & 88.90 & 1.98 & 93.88 & 82.97 & 1.79 & 92.13 & 67.60 & 2.29 & 90.70 & 62.37 & 1.97 \\
  & SliceGPT     & 20\% & 92.40 & 65.27 & 2.63 & 94.08 & 88.23 & 1.56 & 88.48 & 39.93 & 3.25 & 88.80 & 35.00 & 3.40 & 91.98 & 64.13 & 2.27 \\
\bottomrule
\end{tabular}%
\end{table*}

\begin{table*}[hptb]
\centering
\caption{Comparison of Coverage Rate (CR \%), Prediction Accuracy (Acc \%), and Prediction Uncertainty (SS) across other models.}
\scriptsize
\label{tab:compare-main-other-models-combined}
\setlength{\tabcolsep}{2.1pt}
\begin{tabular}{llc ccc ccc ccc ccc ccc}
\toprule 
\multirow{2}{*}{\textbf{Model}} &
  \multirow{2}{*}{\textbf{Method}} &
  \multirow{2}{*}{\textbf{Sparsity}} &
  \multicolumn{3}{c}{\textbf{QA}} &
  \multicolumn{3}{c}{\textbf{RC}} &
  \multicolumn{3}{c}{\textbf{CI}} &
  \multicolumn{3}{c}{\textbf{DRS}} &
  \multicolumn{3}{c}{\textbf{DS}} \\
\cmidrule(lr){4-6} \cmidrule(lr){7-9} \cmidrule(lr){10-12} \cmidrule(lr){13-15} \cmidrule(lr){16-18}
& & & CR & Acc & SS & CR & Acc & SS & CR & Acc & SS & CR & Acc & SS & CR & Acc & SS \\
\midrule
%% ---------- DeepSeek-7B ----------
\multirow{9}{*}{DeepSeek-7B}
  & Dense    & --   & 90.60 & 46.30 & 3.33 & 92.57 & 65.60 & 3.02 & 90.68 & 41.50 & 3.15 & 89.03 & 32.73 & 3.38 & 91.38 & 39.47 & 3.20 \\
\cmidrule(l){2-18}
  & \multicolumn{17}{c}{\textit{Unstructured Pruning}} \\
\cmidrule(l){2-18}
  & Magnitude & 50\% &90.37 & 25.00 & 3.71 & 90.90 & 25.30 & 3.66 & 89.53 & 24.80 & 3.57 & 89.45 & 24.00 & 3.63 & 91.07 & 23.80 & 3.61 \\
  & SparseGPT & 50\% & 91.67 & 38.37 & 3.38 & 89.85 & 49.20 & 3.13 & 88.00 & 26.53 & 3.41 & 88.53 & 32.60 & 3.38 & 91.58 & 24.07 & 3.69 \\
  & Wanda    & 50\% &91.12 & 35.87 & 3.46 & 88.50 & 48.53 & 3.06 & 89.20 & 30.03 & 3.47 & 89.63 & 29.43 & 3.47 & 91.82 & 30.00 & 3.52 \\
\cmidrule(l){2-18}
  & \multicolumn{17}{c}{\textit{Structured Pruning}} \\
\cmidrule(l){2-18}
  & LLM-Pruner   & 20\% & 91.17 & 22.97 & 4.98 & 89.87 & 14.67 & 5.16 & 93.62 & 23.37 & 5.55 & 94.02 & 15.83 & 5.57 & 94.43 & 15.43 & 5.75 \\
  & SliceGPT     & 20\% &  90.63 & 33.33 & 3.55 & 88.95 & 47.80 & 3.06 & 89.65 & 29.87 & 3.55 & 88.35 & 26.87 & 3.50 & 92.17 & 29.93 & 3.71 \\
\midrule
%% ---------- Falcon-7B ----------
\multirow{9}{*}{Falcon-7B}
  & Dense    & --   &  89.55 & 24.53 & 3.76 & 87.98 & 26.50 & 3.48 & 89.67 & 24.10 & 3.59 & 88.27 & 25.70 & 3.53 & 91.30 & 21.30 & 4.40\\
\cmidrule(l){2-18}
  & \multicolumn{17}{c}{\textit{Unstructured Pruning}} \\
\cmidrule(l){2-18}
  & Magnitude& 50\% & 90.88 & 25.00 & 4.72 & 90.13 & 25.43 & 4.62 & 90.43 & 23.50 & 4.64 & 93.53 & 25.70 & 5.64 & 91.10 & 26.40 & 5.48\\
  & SparseGPT & 50\% & 89.60 & 24.80 & 3.75 & 87.92 & 26.57 & 3.49 & 89.75 & 24.03 & 3.59 & 88.27 & 25.70 & 3.53 & 91.27 & 21.27 & 4.40\\
  & Wanda   & 50\% &90.33 & 25.30 & 3.64 & 89.13 & 26.87 & 3.50 & 89.78 & 24.47 & 3.60 & 89.00 & 25.63 & 3.62 & 89.62 & 20.50 & 3.89 \\
\cmidrule(l){2-18}
  & \multicolumn{17}{c}{\textit{Structured Pruning}} \\
\cmidrule(l){2-18}
  & LLM-Pruner   & 20\% & 90.32 & 24.23 & 3.97 & 91.95 & 24.80 & 3.89 & 89.47 & 24.10 & 3.62 & 88.95 & 25.03 & 3.57 & 90.08 & 16.87 & 4.79 \\
  & SliceGPT     & 20\% & 90.75 & 25.57 & 5.21 & 88.60 & 24.20 & 4.33 & 88.85 & 24.10 & 4.30 & 88.47 & 27.57 & 4.16 & 93.42 & 23.93 & 5.31 \\
\bottomrule
\end{tabular}%
\end{table*}

\begin{table*}[hptb]
\centering
\caption{Comparison of Coverage Rate (CR \%), Prediction Accuracy (Acc \%), and Prediction Uncertainty (SS) across Qwen3(MOE) models.}
\scriptsize
\label{tab:compare-main-qwen3-moe-combined}
\setlength{\tabcolsep}{1.6pt}
\begin{tabular}{llc ccc ccc ccc ccc ccc}
\toprule 
\multirow{2}{*}{\textbf{Model}} &
  \multirow{2}{*}{\textbf{Method}} &
  \multirow{2}{*}{\textbf{Sparsity}} &
  \multicolumn{3}{c}{\textbf{QA}} &
  \multicolumn{3}{c}{\textbf{RC}} &
  \multicolumn{3}{c}{\textbf{CI}} &
  \multicolumn{3}{c}{\textbf{DRS}} &
  \multicolumn{3}{c}{\textbf{DS}} \\
\cmidrule(lr){4-6} \cmidrule(lr){7-9} \cmidrule(lr){10-12} \cmidrule(lr){13-15} \cmidrule(lr){16-18}
& & & CR & Acc & SS & CR & Acc & SS & CR & Acc & SS & CR & Acc & SS & CR & Acc & SS \\
\midrule
%% ---------- Qwen3-30B-A3B ----------
\multirow{9}{*}{Qwen3-30B-A3B}
  & Dense    & --   & 93.40 & 66.27 & 3.23 & 92.08 & 64.67 & 2.99 & 91.43 & 36.17 & 3.44 & 89.93 & 25.90 & 4.04 & 91.47 & 39.60 & 2.90 \\
\cmidrule(l){2-18}
  & \multicolumn{17}{c}{\textit{Unstructured Pruning}} \\
\cmidrule(l){2-18}
  & Magnitude & 50\% & 93.12 & 65.00 & 3.13 & 92.12 & 65.23 & 3.15 & 90.87 & 29.73 & 4.18 & 91.28 & 26.63 & 5.30 & 90.63 & 51.90 & 3.28 \\
  & SparseGPT & 50\% & 91.47 & 38.50 & 3.36 & 89.80 & 49.00 & 3.13 & 87.83 & 26.47 & 3.41 & 88.45 & 32.37 & 3.37 & 91.55 & 24.03 & 3.69\\
  & Wanda     & 50\% & 92.80 & 60.17 & 3.62 & 92.08 & 63.83 & 3.85 & 91.87 & 45.53 & 3.23 & 90.92 & 23.60 & 4.35 & 92.65 & 57.23 & 3.15\\
\cmidrule(l){2-18}
  & \multicolumn{17}{c}{\textit{Structured Pruning}} \\
\cmidrule(l){2-18}
  & LLM-Pruner   & 20\% & 90.33 & 76.93 & 2.14 & 90.47 & 83.50 & 1.40 & 91.17 & 83.87 & 1.64 & 89.80 & 62.20 & 2.79 & 90.47 & 65.13 & 2.08  \\
  & SliceGPT     & 20\% & 92.85 & 16.70 & 5.51 & 92.05 & 15.93 & 5.51 & 93.52 & 15.40 & 5.61 & 94.28 & 16.00 & 5.62 & 94.13 & 16.23 & 5.61 \\
\bottomrule
\end{tabular}%
\end{table*}

\begin{table*}[hptb]
\centering
\caption{Comparison of Coverage Rate (CR \%), Prediction Accuracy (Acc \%), and Prediction Uncertainty (SS) across Qwen3(MOE) models.}
\scriptsize
\label{tab:compare-main-qwen-moe}
\setlength{\tabcolsep}{2.1pt}
\begin{tabular}{llc ccc ccc ccc ccc ccc}
\toprule \multirow{2}{*}{\textbf{Model}}
 &
  \multirow{2}{*}{\textbf{Bit}} &
  \multirow{2}{*}{\textbf{Method}} &
  \multicolumn{3}{c}{\textbf{QA}} &
  \multicolumn{3}{c}{\textbf{RC}} &
  \multicolumn{3}{c}{\textbf{CI}} &
  \multicolumn{3}{c}{\textbf{DRS}} &
  \multicolumn{3}{c}{\textbf{DS}} \\
  \cmidrule(lr){4-6} \cmidrule(lr){7-9} \cmidrule(lr){10-12} \cmidrule(lr){13-15} \cmidrule(lr){16-18}
  & & &
  CR & Acc & SS &
  CR & Acc & SS &
  CR & Acc & SS &
  CR & Acc & SS &
  CR & Acc & SS \\
\midrule
\multirow{4}{*}{Qwen3-30B-A3B}
  & FP16   & --        & 93.40 & 66.27 & 3.23 & 92.08 & 64.67 & 2.99 & 91.43 & 36.17 & 3.44 & 89.93 & 25.90 & 4.04 & 91.47 & 39.60 & 2.90\\
  \noalign{\vspace{0.1em}}\cdashline{2-18}\noalign{\vspace{0.1em}}
  & W4A16  & RTN       &94.03 & 64.90 & 3.05 & 92.77 & 63.40 & 2.28 & 91.70 & 35.20 & 3.28 & 89.97 & 25.10 & 3.88 & 92.08 & 39.60 & 2.98 \\
  & W4A16  & AWQ       &91.87 & 52.07 & 3.51 & 90.83 & 58.43 & 2.79 & 91.52 & 33.33 & 3.49 & 88.07 & 20.65 & 4.30 & 91.07 & 21.38 & 3.84  \\
  & W4A16  & GPTQ      &92.82&67.07&3.04&93.68&76.03&2.45&90.68&39.20&3.38&89.81&34.37&3.86&12.68&25.35&1.00\\
\midrule
\end{tabular}%
\end{table*}

\begin{table*}[hptb]
\centering
\label{tab:llama2-progressive}
\caption{Comparison of Coverage Rate (CR \%), Prediction Accuracy (Acc \%), and Prediction Uncertainty (SS) across Llama2 models.}
\scriptsize
\label{tab:llama2-progressive}
\setlength{\tabcolsep}{2.4pt}
\begin{tabular}{llc ccc ccc ccc ccc ccc}
\toprule 
\multirow{2}{*}{\textbf{Model}} &
  \multirow{2}{*}{\textbf{Method}} &
  \multirow{2}{*}{\textbf{Sparsity}} &
  \multicolumn{3}{c}{\textbf{QA}} &
  \multicolumn{3}{c}{\textbf{RC}} &
  \multicolumn{3}{c}{\textbf{CI}} &
  \multicolumn{3}{c}{\textbf{DRS}} &
  \multicolumn{3}{c}{\textbf{DS}} \\
\cmidrule(lr){4-6} \cmidrule(lr){7-9} \cmidrule(lr){10-12} \cmidrule(lr){13-15} \cmidrule(lr){16-18}
& & & CR & Acc & SS & CR & Acc & SS & CR & Acc & SS & CR & Acc & SS & CR & Acc & SS \\
\midrule
%% ---------- Llama2-7B ----------
\multirow{9}{*}{Llama2-7B}
  & Dense    & --   & 91.40 & 45.53 & 3.20 & 91.83 & 67.20 & 2.46 & 90.47 & 43.27 & 3.26 & 90.73 & 32.67 & 3.30 & 89.85 & 46.77 & 3.24\\
\cmidrule(l){2-18}
  & wanda    & 10\% & 91.40 & 45.73 & 3.23 & 91.92 & 67.10 & 2.51 & 90.47 & 41.37 & 3.32 & 91.05 & 32.87 & 3.38 & 89.53 & 46.43 & 3.35\\
\cmidrule(l){2-18}
  & wanda    & 20\% &  90.88 & 45.30 & 3.33 & 91.35 & 66.70 & 2.52 & 91.37 & 42.07 & 3.36 & 91.45 & 32.70 & 3.47 & 89.82 & 44.67 & 3.39 \\
\cmidrule(l){2-18}
  & wanda    & 30\% &  90.60 & 43.87 & 3.46 & 91.75 & 63.63 & 2.73 & 91.05 & 41.00 & 3.37 & 90.33 & 32.47 & 3.51 & 90.72 & 46.20 & 3.27 \\
\cmidrule(l){2-18}
  & wanda    & 40\% & 90.75 & 41.57 & 3.72 & 89.93 & 52.83 & 2.89 & 90.43 & 31.90 & 3.44 & 89.30 & 29.30 & 3.55 & 90.97 & 31.80 & 3.21 \\
\cmidrule(l){2-18}
  & Wanda    & 50\% &   89.40 & 35.57 & 3.59 & 91.25 & 37.60 & 3.48 & 91.03 & 26.57 & 3.77 & 89.30 & 27.07 & 3.61 & 90.15 & 28.83 & 3.45\\
\midrule
%% ---------- Llama2-13B ----------
\multirow{9}{*}{Llama2-13B}
  & Dense    & --   & 92.47 & 53.10 & 3.10 & 94.45 & 77.43 & 2.33 & 92.75 & 59.63 & 2.83 & 90.77 & 53.17 & 2.57 & 90.00 & 60.27 & 2.15 \\
\cmidrule(l){2-18}
  & wanda    & 10\% &  92.55 & 52.97 & 3.12 & 94.38 & 77.27 & 2.35 & 92.88 & 59.97 & 2.80 & 90.92 & 53.97 & 2.56 & 90.32 & 62.37 & 2.12\\
\cmidrule(l){2-18}
  & wanda    & 20\% &  92.28 & 51.83 & 3.13 & 94.35 & 78.07 & 2.36 & 92.97 & 60.37 & 2.88 & 90.55 & 53.00 & 2.69 & 90.65 & 63.03 & 2.07 \\
\cmidrule(l){2-18}
  & wanda    & 30\% & 92.35 & 50.70 & 3.05 & 94.62 & 77.70 & 2.31 & 92.23 & 56.50 & 2.95 & 89.73 & 50.73 & 2.71 & 91.25 & 65.07 & 1.98 \\
\cmidrule(l){2-18}
  & wanda    & 40\% &  91.33 & 49.17 & 3.08 & 93.32 & 74.73 & 2.23 & 91.17 & 51.77 & 3.07 & 89.00 & 48.20 & 2.64 & 91.07 & 59.17 & 2.17 \\
\cmidrule(l){2-18}
  & Wanda    & 50\% & 91.77 & 45.57 & 3.20 & 92.50 & 66.20 & 2.56 & 89.20 & 39.83 & 3.46 & 89.73 & 40.83 & 3.21 & 90.55 & 53.10 & 2.44  \\
\midrule
%% ---------- Llama2-70B ----------
\multirow{9}{*}{Llama2-70B}
  & Dense    & --   & 93.23 & 65.63 & 2.64 & 95.83 & 90.80 & 1.79 & 94.45 & 82.37 & 1.85 & 91.93 & 67.40 & 2.33 & 90.02 & 56.20 & 2.23\\
\cmidrule(l){2-18}
  & wanda    & 10\% & 93.45 & 65.30 & 2.63 & 95.67 & 90.43 & 1.79 & 94.08 & 82.77 & 1.85 & 92.03 & 67.27 & 2.34 & 89.68 & 56.10 & 2.23\\
\cmidrule(l){2-18}
  & wanda    & 20\% &  93.32 & 64.73 & 2.66 & 95.80 & 90.33 & 1.84 & 93.88 & 81.73 & 1.85 & 91.60 & 67.73 & 2.32 & 89.62 & 56.80 & 2.21 \\
\cmidrule(l){2-18}
  & wanda    & 30\% &  93.17 & 64.53 & 2.69 & 95.82 & 90.23 & 1.89 & 93.77 & 81.87 & 1.83 & 90.78 & 66.00 & 2.27 & 89.75 & 56.07 & 2.20 \\
\cmidrule(l){2-18}
  & wanda    & 40\% &  92.53 & 63.73 & 2.76 & 95.67 & 89.13 & 1.94 & 93.65 & 79.77 & 1.94 & 91.50 & 64.20 & 2.45 & 90.37 & 52.43 & 2.43 \\
\cmidrule(l){2-18}
  & Wanda    & 50\% & 91.95 & 60.53 & 2.85 & 95.35 & 87.97 & 1.94 & 92.75 & 72.57 & 2.14 & 90.45 & 59.53 & 2.60 & 91.13 & 52.53 & 2.55  \\
\bottomrule
\end{tabular}%
\end{table*}

\begin{table*}[hptb]
\centering
\caption{Comparison of Coverage Rate (CR \%), Prediction Accuracy (Acc \%), and Prediction Uncertainty (SS) across Llama2 models.}
\scriptsize
\label{tab:llama3-progressive}
\setlength{\tabcolsep}{2.4pt}
\begin{tabular}{llc ccc ccc ccc ccc ccc}
\toprule 
\multirow{2}{*}{\textbf{Model}} &
  \multirow{2}{*}{\textbf{Method}} &
  \multirow{2}{*}{\textbf{Sparsity}} &
  \multicolumn{3}{c}{\textbf{QA}} &
  \multicolumn{3}{c}{\textbf{RC}} &
  \multicolumn{3}{c}{\textbf{CI}} &
  \multicolumn{3}{c}{\textbf{DRS}} &
  \multicolumn{3}{c}{\textbf{DS}} \\
\cmidrule(lr){4-6} \cmidrule(lr){7-9} \cmidrule(lr){10-12} \cmidrule(lr){13-15} \cmidrule(lr){16-18}
& & & CR & Acc & SS & CR & Acc & SS & CR & Acc & SS & CR & Acc & SS & CR & Acc & SS \\
\midrule
%% ---------- Llama3.2-1B ----------
\multirow{9}{*}{Llama3.2-1B}
  & Dense    & --   & 92.02 & 33.87 & 3.55 & 90.12 & 38.27 & 3.43 & 88.82 & 24.13 & 3.69 & 88.50 & 24.67 & 3.54 & 92.03 & 24.37 & 3.62\\
\cmidrule(l){2-18}
  & wanda    & 10\% &  91.85 & 32.67 & 3.53 & 90.35 & 38.00 & 3.44 & 88.95 & 24.20 & 3.69 & 88.67 & 24.97 & 3.55 & 92.37 & 23.57 & 3.67 \\
\cmidrule(l){2-18}
  & wanda    & 20\% &  91.32 & 31.17 & 3.56 & 89.57 & 33.07 & 3.46 & 88.70 & 24.10 & 3.67 & 88.93 & 24.73 & 3.55 & 91.75 & 24.23 & 3.69 \\
\cmidrule(l){2-18}
  & wanda    & 30\% & 90.53 & 28.87 & 3.54 & 90.07 & 30.90 & 3.52 & 88.82 & 24.10 & 3.69 & 89.40 & 26.23 & 3.56 & 91.65 & 24.07 & 3.65 \\
\cmidrule(l){2-18}
  & wanda    & 40\% &  89.32 & 27.10 & 3.67 & 89.27 & 26.63 & 3.56 & 90.12 & 25.13 & 3.83 & 89.33 & 25.87 & 3.59 & 89.13 & 13.80 & 4.71 \\
\cmidrule(l){2-18}
  & Wanda    & 50\% & 89.48 & 24.67 & 3.60 & 89.08 & 25.47 & 3.56 & 89.65 & 24.70 & 3.61 & 89.27 & 25.70 & 3.67 & 91.15 & 25.87 & 3.81\\
\midrule
%% ---------- Llama3.1-8B ----------
\multirow{9}{*}{Llama3.1-8B}
  & Dense   & --   &  91.72 & 62.93 & 2.98 & 94.85 & 87.07 & 1.89 & 92.15 & 57.60 & 2.80 & 92.57 & 63.30 & 2.68 & 90.20 & 60.47 & 1.89  \\
\cmidrule(l){2-18}
  & Wanda  & 10\% &  91.98 & 63.53 & 3.01 & 94.70 & 86.50 & 1.91 & 91.80 & 55.80 & 2.76 & 92.57 & 63.53 & 2.75 & 89.87 & 59.63 & 1.89\\
\cmidrule(l){2-18}
  & Wanda   & 20\% &91.38 & 63.10 & 3.14 & 94.67 & 86.53 & 1.94 & 90.80 & 57.87 & 2.75 & 92.63 & 62.60 & 2.75 & 90.23 & 59.50 & 1.95\\
\cmidrule(l){2-18}
  & Wanda   & 30\% & 92.23 & 61.40 & 2.95 & 93.80 & 83.70 & 2.01 & 89.38 & 59.60 & 2.63 & 92.80 & 58.53 & 2.66 & 90.90 & 55.90 & 2.04 \\
\cmidrule(l){2-18}
  & Wanda   & 40\% & 93.07 & 57.40 & 3.00 & 94.07 & 81.00 & 2.06 & 91.03 & 52.07 & 3.04 & 91.75 & 51.20 & 2.54 & 89.68 & 48.53 & 2.09\\
\cmidrule(l){2-18}
  & Wanda   & 50\% &90.73 & 45.97 & 3.09 & 90.40 & 59.40 & 2.53 & 90.65 & 31.67 & 3.42 & 89.58 & 35.50 & 3.16 & 89.90 & 45.97 & 2.52\\
\midrule
%% ---------- Llama3.1-70B ----------
\multirow{9}{*}{Llama3.1-70B}
  & Dense    & --   & 92.53 & 76.20 & 2.17 & 96.32 & 93.13 & 1.61 & 94.43 & 81.33 & 1.89 & 91.23 & 74.97 & 1.88 & 90.95 & 62.37 & 2.06 \\
\cmidrule(l){2-18}
  & wanda    & 10\% &  92.82 & 76.10 & 2.17 & 96.23 & 93.00 & 1.64 & 94.38 & 81.13 & 1.90 & 90.92 & 73.57 & 1.92 & 90.93 & 61.03 & 2.20\\
\cmidrule(l){2-18}
  & wanda    & 20\% & 92.60 & 74.93 & 2.19 & 95.87 & 92.50 & 1.64 & 94.50 & 78.73 & 2.04 & 90.78 & 72.80 & 1.97 & 91.58 & 59.20 & 2.43 \\
\cmidrule(l){2-18}
  & wanda    & 30\% & 92.97 & 75.03 & 2.20 & 96.30 & 92.83 & 1.72 & 94.48 & 76.23 & 2.21 & 91.78 & 71.10 & 2.04 & 91.05 & 57.13 & 2.29 \\
\cmidrule(l){2-18}
  & wanda    & 40\% & 93.58 & 74.37 & 2.28 & 95.40 & 90.73 & 1.66 & 91.30 & 60.17 & 2.66 & 91.47 & 69.40 & 2.01 & 91.30 & 61.57 & 1.99 \\
\cmidrule(l){2-18}
  & Wanda    & 50\% & 93.65 & 71.37 & 2.33 & 94.33 & 88.83 & 1.56 & 90.80 & 49.53 & 3.16 & 89.05 & 58.80 & 2.32 & 90.88 & 55.97 & 2.07 \\
\bottomrule
\end{tabular}%
\end{table*}

\begin{table*}[hptb]
\centering
\caption{Comparison of Coverage Rate (CR \%), Prediction Accuracy (Acc \%), and Prediction Uncertainty (SS) across Qwen3(Dense) models.}
\scriptsize
\label{tab:compare-main-qwen-dense}
\setlength{\tabcolsep}{2.4pt}
\begin{tabular}{llc ccc ccc ccc ccc ccc}
\toprule \multirow{2}{*}{\textbf{Model}}
 &
  \multirow{2}{*}{\textbf{Bit}} &
  \multirow{2}{*}{\textbf{Method}} &
  \multicolumn{3}{c}{\textbf{QA}} &
  \multicolumn{3}{c}{\textbf{RC}} &
  \multicolumn{3}{c}{\textbf{CI}} &
  \multicolumn{3}{c}{\textbf{DRS}} &
  \multicolumn{3}{c}{\textbf{DS}} \\
  \cmidrule(lr){4-6} \cmidrule(lr){7-9} \cmidrule(lr){10-12} \cmidrule(lr){13-15} \cmidrule(lr){16-18}
  & & &
  CR & Acc & SS &
  CR & Acc & SS &
  CR & Acc & SS &
  CR & Acc & SS &
  CR & Acc & SS \\
\midrule
\multirow{7}{*}{Qwen3-8B}
  & FP16   & --        &94.20 & 70.43 & 3.00 & 93.70 & 84.17 & 2.20 & 94.67 & 82.10 & 2.31 & 90.40 & 58.03 & 3.35 & 90.57 & 57.47 & 2.56 \\
  \noalign{\vspace{0.1em}}\cdashline{2-18}\noalign{\vspace{0.1em}}
  & W4A16  & RTN       &93.24 & 70.76 & 2.97 & 93.27 & 80.44 & 2.43 & 93.66 & 79.57 & 2.61 & 90.34 & 54.17 & 3.46 & 90.16 & 50.94 & 3.14 \\
  & W4A16  & AWQ       &94.43 & 72.03 & 3.11 & 93.85 & 83.00 & 2.27 & 94.03 & 80.47 & 2.31 & 92.33 & 57.12 & 3.07 & 90.48 & 55.41 & 2.64  \\
  & W4A16  & GPTQ      &93.37 & 71.93 & 2.86 & 93.63 & 81.67 & 2.39 & 94.04 & 80.93 & 2.48 & 90.57 & 55.33 & 3.39 & 90.53 & 51.77 & 3.03 \\
  \noalign{\vspace{0.1em}}\cdashline{2-18}\noalign{\vspace{0.1em}}
  & W4A4   & FlatQuant          & 91.00&17.20&5.37&89.87&19.70&4.99&91.27&17.73&5.26&89.64&18.22&5.05&91.58&15.20&5.20
\\
  & W4A4   & QuaRot    &92.67&14.46&5.51&92.14&18.67&5.44&91.77&18.88&5.50&93.04&19.01&5.50&91.37&20.28&5.36\\
  & W4A4   & SpinQuant & -- & -- & -- & -- & -- &  -- & -- & -- & -- & -- & -- & -- & -- & -- & -- \\
\midrule
\multirow{7}{*}{Qwen3-14B}
  & FP16   & --        &94.02 & 74.33 & 2.71 & 94.38 & 84.97 & 2.09 & 94.83 & 86.60 & 1.86 & 91.12 & 56.10 & 3.16 & 89.48 & 60.50 & 2.25 \\
  \noalign{\vspace{0.1em}}\cdashline{2-18}\noalign{\vspace{0.1em}}
  & W4A16  & RTN       &93.23 & 71.80 & 2.79 & 94.28 & 84.07 & 2.14 & 94.03 & 84.70 & 1.94 & 89.52 & 56.22 & 3.26 & 91.08 & 60.75 & 2.58  \\
  & W4A16  & AWQ       &93.90 & 72.63 & 2.82 & 94.42 & 82.37 & 2.17 & 94.87 & 85.37 & 2.00 & 92.79 & 62.26 & 2.95 & 91.77 & 60.99 & 2.42\\
  & W4A16  & GPTQ      & 94.05 & 74.30 & 2.77 & 93.97 & 82.87 & 2.17 & 94.87 & 84.67 & 2.05 & 91.71 & 58.93 & 3.03 & 91.55 & 59.99 & 2.56 \\
  \noalign{\vspace{0.1em}}\cdashline{2-18}\noalign{\vspace{0.1em}}
  & W4A4   &  FlatQuant         &94.00&19.93&5.59&89.90&22.53&5.37&92.05&21.17&5.49&89.57&19.25&5.38&93.64&21.61&5.58
 \\
  & W4A4   & QuaRot    &93.37 & 13.43 & 5.70 & 92.10 & 18.53 & 5.47 & 93.63 & 11.67 & 5.73 & 92.23 & 18.42 & 5.69 & 92.79 & 18.54 & 5.44 \\
  & W4A4   & SpinQuant & -- & -- & -- & -- & -- &  -- & -- & -- & -- & -- & -- & -- & -- & -- & -- \\
\midrule
\multirow{7}{*}{Qwen3-32B}
  & FP16   & --        &65.18 & 58.63 & 2.06 & 93.23 & 83.97 & 1.93 & 93.83 & 86.97 & 1.73 & 90.95 & 62.03 & 3.31 & 80.90 & 64.03 & 1.79\\
  \noalign{\vspace{0.1em}}\cdashline{2-18}\noalign{\vspace{0.1em}}
  & W4A16  & RTN       &65.25 & 58.80 & 2.08 & 92.60 & 82.40 & 1.98 & 93.80 & 86.40 & 1.78 & 89.70 & 62.40 & 3.10 & 86.50 & 61.20 & 2.10\\
  & W4A16  & AWQ       &65.05 & 58.30 & 2.08 & 93.88 & 83.83 & 1.93 & 94.78 & 87.27 & 1.82 & 90.19 & 65.20 & 2.92 & 71.96 & 62.16 & 1.39 \\
  & W4A16  & GPTQ      &65.45 & 58.03 & 2.17 & 92.58 & 80.03 & 2.05 & 94.12 & 85.20 & 1.90 & 90.32 & 59.26 & 3.57 & 82.26 & 63.69 & 1.93 \\
  \noalign{\vspace{0.1em}}\cdashline{2-18}\noalign{\vspace{0.1em}}
  & W4A4   &  FlatQuant         &90.50 & 20.97 & 4.51 & 90.33 & 25.43 & 4.08 & 90.27 & 21.43 & 4.35 & 90.16 & 22.62 & 4.00 & 88.99 & 21.81 & 4.25\\
  & W4A4   & QuaRot    &91.00 & 21.23 & 5.40 & 90.78 & 21.83 & 5.25 & 90.72 & 20.23 & 5.33 & 89.86 & 21.99 & 5.08 & 90.55 & 17.17 & 5.22 \\
  & W4A4   & SpinQuant & -- & -- & -- & -- & -- &  -- & -- & -- & -- & -- & -- & -- & -- & -- & --  \\
\bottomrule
\end{tabular}%
\end{table*}

\end{document}